\documentclass{article}

\usepackage[preprint]{neurips_2026}

\usepackage[utf8]{inputenc} 
\usepackage[T1]{fontenc}    
\usepackage{hyperref}       
\usepackage{url}            
\usepackage{booktabs}       
\usepackage{amsfonts}       
\usepackage{nicefrac}       
\usepackage{microtype}      
\usepackage[dvipsnames]{xcolor}         
\usepackage{hyperref}
\usepackage{url}
\usepackage{enumitem}
\usepackage{amsmath}
\usepackage{multirow}
\usepackage{graphicx}
\usepackage[most]{tcolorbox}
\usepackage{tcolorbox}
\tcbuselibrary{breakable}
\usepackage{wrapfig}
\usepackage{subcaption}

\title{PermuFormer: Multi-Task Pretraining for Permutation Representation in Algebraic Combinatorics}

\author{%
  Henry Kvinge \\
  Pacific Northwest National Laboratory \\
  University of Washington \\
  \texttt{henry.kvinge@pnnl.gov} \\
}

\begin{document}

\maketitle

\begin{abstract}
Diverse pretraining has been shown to be an effective method for learning reusable, domain-aware representations that provide a starting point for fine-tuning on downstream tasks. While much of the excitement in AI for math has been concentrated in the use of frontier reasoning models to solve well-specified problems through the medium of language, narrow, specialized models remain an important component of the AI for math ecosystem. In contrast to large language models, specialized models are usually trained directly on the mathematical objects themselves (e.g., graphs, sequences of numbers) rather than the textual descriptions that characterize these objects. However, the common practice of training specialists from scratch may limit their ability to develop domain-aware representations that capture the multifaceted nature of mathematics. In this paper, we describe an approach to pretraining for permutation-focused tasks in algebraic combinatorics. We introduce PermuFormer, an autoregressive transformer trained on a 2.8 billion token multi-task, multi-encoding corpus. We show that PermuFormer is an effective starting point for fine-tuning on basic tasks unseen during pretraining and more complex research-level tasks, frequently outperforming the same architecture trained from scratch, baseline MLPs, and a fine-tuned generic language model of comparable size. We also analyze some of the internal mechanisms by which PermuFormer learns to solve training tasks. For example, we show that while some tasks can be linearly decoded directly from the internal representation of the prompt, other tasks require multiple rounds of generation before the answer can be decoded.
\end{abstract}


\begin{figure}[htbp]
    \centering
    \includegraphics[width=0.9\textwidth]{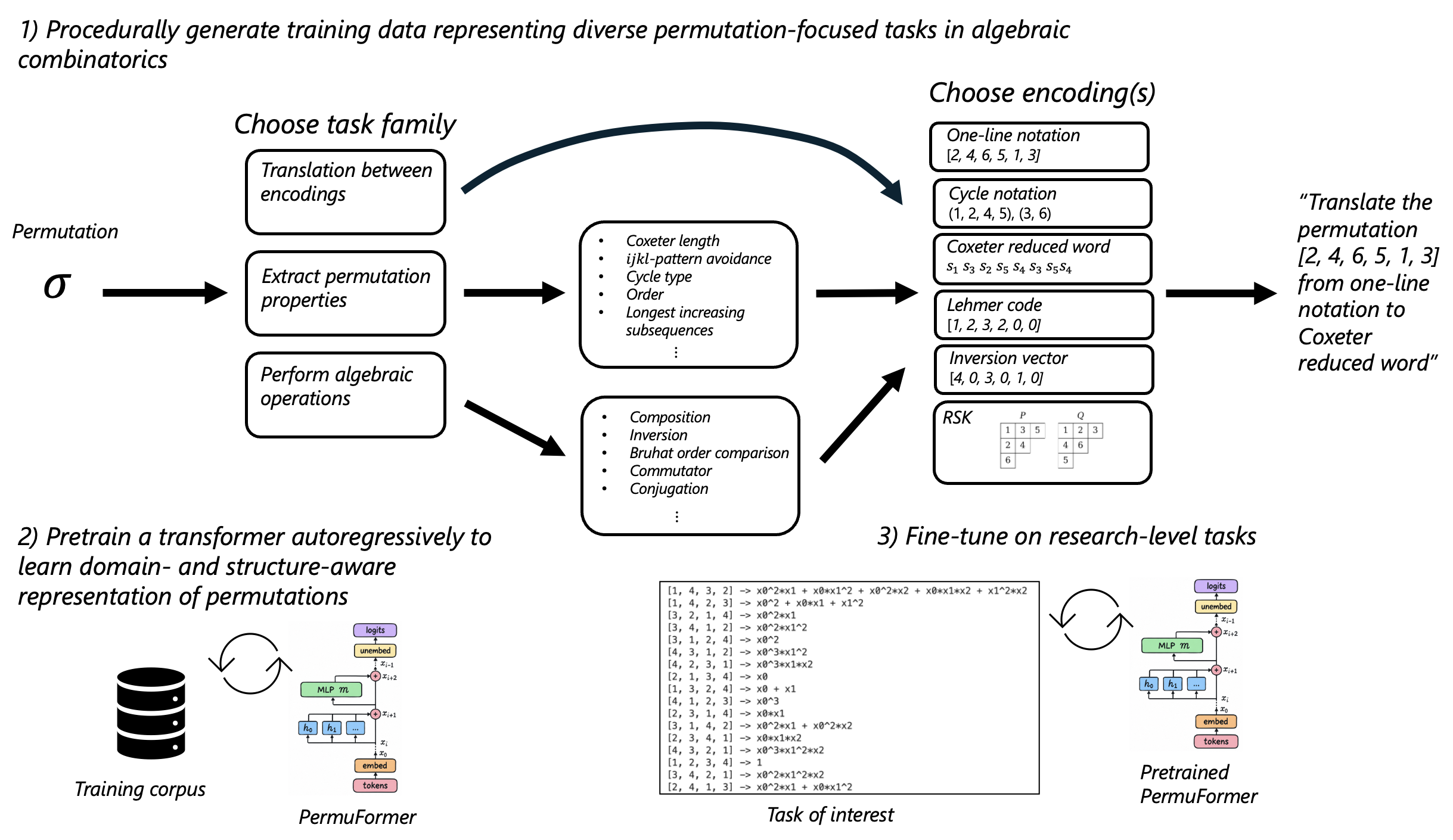}
    \caption{We generate data representing diverse, permutation-focused tasks in algebraic combinatorics using six different ways of encoding permutations. We then pretrain an autoregressive transformer, PermuFormer, on this data. PermuFormer learns reusable features that make it an effective starting point for fine-tuning on downstream tasks in algebraic combinatorics.}
    \label{fig:permuformer-diagram}
\end{figure}

\section{Introduction}

The concept of pretraining a model on diverse data, broader than what is required for the downstream task, is a central approach to obtaining robust and transferable representations in deep learning models. Pretraining now plays an essential role in a range of modalities and application areas including language \cite{devlin-etal-2019-bert}, image generation \cite{Rombach_2022_CVPR}, speech-to-text \cite{baevski2020wav2vec}, and image recognition \cite{simclr} and detection \cite{Girshick_2014_CVPR}. In the case of language models, for instance, pretraining provides a model with an understanding of language along with exposure to a vast amount of information encoded in that language. For any modality, pretraining helps a model learn the structure of the modality, including reusable and robust features that can then be modified for specific downstream tasks. Notably, pretraining is now also a fundamental step in the development of neural networks for scientific tasks \cite{hu2019strategies,climax}. This is especially interesting since scientific tasks often involve data for which human lack intuition for.

While small specialist models remain a core component of the AI for math ecosystem, they are typically trained from scratch on a specific task of interest. Capturing the multifaceted nature of mathematics may be challenging when one restricts to this narrow approach. To try to address this limitation, in this work we explore pretraining for tasks around permutations. Permutations are among the central objects of study in combinatorics and are a fundamental tool across other areas of math and the sciences. In mathematics, properties of permutations often govern the behavior of seemingly unrelated structures such as the geometry of Grassmann manifolds or the algebraic structure of Lie groups \cite{fulton1997young,manivel2001symmetric,sagan2001symmetric}. Permutations are also a core ingredient in many fields of physics \cite{baxter1985exactly,baez1994gauge, zee2016group} and computer science \cite{knuth1997art,knuth1998art,knuth2013art,knuth2014art}. Within algebraic combinatorics, permutations are integrated into a web of related objects like integer partitions, graphs, and Young tableaux that together express universal combinatorial patterns. While significant effort has been put into developing machine learning models capable of extracting rich representations from some of these structures in isolation (such as graphs \cite{liu2023towards}), permutations have received far less attention. 

To address this, we introduce \emph{PermuFormer}\footnote{PermuFormer can be found at \url{https://huggingface.co/ACDRepo/PermuFormer}, its training and evaluation sets can be found at \url{https://huggingface.co/datasets/ACDRepo/permuformer_data}}, a transformer trained to perform permutation-centric tasks from combinatorics. We design a 2.8-billion-token multi-task, multi-encoding\footnote{Because the word `representation' is already heavily used in machine learning, we use the word \emph{encoding} to denote the form in which one writes a permutation and \emph{representation} to denote the internal activations of a deep learning model.} training set that captures the breadth of roles that permutations play in combinatorics. A single task is an operation performed on one or more permutations encoded in different ways. Encodings include one-line notation, cycle notation, and reduced Coxeter expression. Tasks range from algebraic operations on permutations (composition, inversion, and conjugation), to the extraction of statistics like length, pattern avoidance, and fixed points, to translation between different encodings of permutations. Through the diversity of this training regime, PermuFormer begins to capture the rich, interconnected, and multifaceted landscape of permutations in combinatorics. This approach, which trains a small, 75-million-parameter model on many permutation calculations, complements the language-focused and compute-heavy reasoning model paradigm represented by recent successes in AI for mathematics. 

\textbf{What is the value of PermuFormer?} Because PermuFormer is pretrained across a diverse set of tasks, it is an effective starting point for machine learning-based exploration of research-level problems in combinatorics. We demonstrate this via our fine-tuning experiments (Section \ref{sect:fine-tuning}) which include, standard tasks like pattern avoidance as well as research-level problems such as the prediction of Schubert polynomial structure constants and the coefficients and degrees of Kazhdan-Lusztig polynomials. We show that when PermuFormer is fine-tuned on a downstream task, it usually outperforms an identical architecture trained from scratch, baseline MLPs, and a fine-tuned generic language model of comparable size. This suggests that PermuFormer has learned reusable features for permutations that are valuable for many tasks in algebraic combinatorics.

Beyond just being useful for fine-tuning on downstream research tasks, PermuFormer may also provide a setting for studying how a transformer learns a complex web of interrelated mathematical tasks from data alone. We believe this is an especially interesting angle given that PermuFormer learns to solve tasks using many different encodings of permutations. We show, for instance, that the layer and generation step at which we can linearly decode a permutation property from hidden representations is strongly dependent on both the task and the permutation encoding. However, on certain tasks, the hidden representations corresponding to two distinct encodings may become highly similar when compared via centered kernel alignment (CKA) \cite{kornblith2019similarity}. Together, this suggests that PermuFormer adopts a flexible strategy for representing permutations, keeping encoding-specific features in the context of certain tasks and discarding them for others. We also provide an initial analysis of the global geometry of permutations in the context of different tasks and encodings, suggesting that while local geometry is relatively stable across encodings, the global geometry can vary substantially.

In summary, the contributions of this paper are the following:
\begin{itemize}[leftmargin=1.5em, itemsep=0.25em]
    \item We introduce a 2.8 billion token dataset spanning 306 permutation-focused tasks designed for pretraining in algebraic combinatorics.
    \item We train an autoregressive transformer, PermuFormer, that achieves high accuracy on these tasks by learning rich representations that transfer to a range of research-level downstream tasks via fine-tuning.
    \item We provide an analysis of how PermuFormer represents permutations across encodings, tasks, and generation steps, shedding light on how transformers learn to represent fundamental mathematical objects.
\end{itemize}

\subsection{Related Work}

\textbf{Pretraining for mathematical domains:} The most well-known cases of pretraining for mathematics involve generalist frontier models like the  GPT-o1, GPT-o3, and GPT-o4 series \cite{jaech2024openai}, DeepSeek-R1 \cite{guo2025deepseek}, Fable \cite{anthropic2026fable}, and more recent GPT models like Sol and Astra \cite{openai2026sol,openai2026astra}. These models are trained on internet-scale data. This may include many mathematics papers and textbooks but it also includes a vast number of documents representing other domains. 

Other models have been more narrowly targeted at math such as AlphaGeometry \cite{trinh2024solving,chervonyi2025gold}, an AI system augmented with a geometry-specialized symbolic engine. Like PermuFormer, this model was pretrained on synthetically generated data, but AlphaGeometry's pretraining data consists of geometry proofs rather than direct computations. Other neurosymbolic systems that combine learned models with proof assistants include \cite{lin2025goedel,yang2023leandojo,dong2025stp,hubert2025olympiad,chervonyi2025gold}. Like AlphaGeometry, these are trained for the task of generating proofs via formal systems rather than direct computation on mathematical objects.

Perhaps most similar to the present work are machine learning models that use broad pretraining directly on structured mathematical objects. Important examples include graph foundation models \cite{bechler2026billion,yu2026riemannian,wang2025graph,wang2024towards}, models pretrained on the Online Encyclopedia of Integer Sequences \cite{nextterm47m2025,nextterm440m2026,nakasho2026intseqbert}, and specific pretraining on arithmetic for numeracy \cite{petrak-etal-2023-arithmetic}. To our knowledge, PermuFormer is the first example of multi-task pretraining specifically for permutation representations.

\textbf{Machine learning and algebraic combinatorics:} Given that algebraic combinatorics studies discrete, computationally accessible structures like graphs, integer partitions, and permutations, it is a natural candidate for machine learning. A prominent example of this is one of the two case studies presented in \cite{davies2021advancing}, which used a graph neural network to study intervals in the Bruhat graph related to the combinatorial invariance conjecture. Other notable works include \cite{chau2025machine}, which introduced several datasets representing either open or foundational problems in algebraic combinatorics as well as initial benchmarking using both narrow models and LLMs. In a different direction, several works have explored the use of machine learning to find bijections between combinatorial objects \cite{huang2026discovering,brown2025even}. Closely related to our PermuFormer work is \cite{petschack2025learning}, which showed that transformers are capable of generalizing computation learned from small symmetric groups to larger symmetric groups. We note that this type of generalization is complementary to the aim of the present work, which looks at generalization to new tasks using a fixed range of symmetric groups. \cite{petschack2025learning} also does not study pretraining across many different tasks or permutation encodings. Finally, there are a range of other works that use transformers to study combinatorics-adjacent problems from fields such as enumerative geometry \cite{hashemi2025can}.

\textbf{Synthetic testbeds for language models:} The idea of using simple synthetic systems to model how language models learn and process the more complex real world has long held appeal. Board games have been a particularly rich source of such small worlds \cite{toshniwal2022chess,li2022emergent}. Other works have looked at the use of simple algebraic frameworks like modular arithmetic or small symmetric groups to understand the underlying mechanisms of model computation \cite{nanda2023progress,stander2023grokking} or to understand the extent to which the model has internalized general mathematical rules \cite{kvingecan}. In many cases, such studies have resulted in new insights into the similarities and differences between learned representations and the underlying structures that they model \cite{vafa2024evaluating,vafa2025has}. \cite{scullen2025permutations} is most closely related to our representational analysis. This paper advocates for the use of permutations and their encodings as a setting to study the relationship between input encoding and learning dynamics. Unlike that work, which trained different models for each permutation-encoding/task pair, we train a single, larger model across all encodings and tasks.

\section{Permutations, their Encodings, and their Properties}
\label{sect-background}

Permutations are our central object of study in this work. A permutation $\sigma$ on $n$ elements is a bijective map from $\{1,2,\dots,n\}$ to itself. Henceforth, we denote $[n] := \{1,2,\dots,n\}$. 

The set of $n!$ permutations of $[n]$ forms a group called the \emph{symmetric group} $S_n$. The symmetric group's binary operation is composition of permutations (as functions). The permutation that swaps the two adjacent elements $i,i+1$ and fixes all others is denoted by $s_i$ and is known as a \emph{Coxeter generator}. It plays a special role in the study of permutations. The word `generator' comes from the fact that any permutation in $S_n$ can be generated as a composition of these elements.  We use the convention that elements of the symmetric group act on integers in $[n]$ on the left, and this defines their composition. As a running example, we use the permutation $\tau \in S_5$ that sends $1 \mapsto 3, 2 \mapsto 1, 3 \mapsto 2, 4 \mapsto 5, 5 \mapsto 4$. This can be written as a composition of Coxeter generators as $\tau = s_2s_1s_4$. 

An \emph{inversion}\footnote{Not to be confused with the algebraic inverse of a permutation, a permutation $\sigma^{-1}$ such that $\sigma \sigma^{-1} = 1$.} in a permutation $\sigma \in S_n$ is a pair $(i,j)$ such that $i < j$ but $\sigma(i) > \sigma(j)$. The number of inversions in a permutation $\sigma$ is called the \emph{Coxeter length} of $\sigma$. This statistic can also be defined as the minimal number of Coxeter generators required to generate $\sigma$. In the example $\tau$ above, we have three inversions, $(1,2)$, $(1,3)$, and $(4,5)$. This means that $\tau$ has length 3. The \emph{sign} of a permutation $\sigma$ can be defined as
\begin{equation*}
    \operatorname{sgn}(\sigma) = (-1)^{\text{length}(\sigma)}.
\end{equation*}
If $\operatorname{sgn}(\sigma)=1$, then $\sigma$ is called \emph{even}. If $\operatorname{sgn}$ is $-1$, $\sigma$ is called \emph{odd}. $\tau$ is an odd permutation since $(-1)^3 = -1$.

Recall that an integer partition $\lambda$ of $n$ is a weakly decreasing sequence of integers $\lambda = (n_1,n_2,\dots,n_l)$ such that $n_1 + n_2 + \dots + n_l = n$. We can realize a partition $\lambda = (n_1,\dots,n_l)$ of $n$ as a \emph{Young diagram} which is a collection of $n$ boxes, arranged in $l$ left-justified rows with $n_1$ boxes in the first row, $n_2$ boxes in the second row, etc. A \emph{standard Young tableau} of shape $\lambda$ is a filling of the boxes in the Young diagram corresponding to $\lambda$ with elements from $[n]$ such that entries increase left to right across rows and down columns.

The \emph{Robinson-Schensted-Knuth (RSK) algorithm} associates each permutation with a pair of standard Young tableaux of the same shape \cite{stanley2011enumerative}. This bijection is central to the field since it gives a correspondence between two of the fundamental data types in combinatorics: permutations and standard Young tableaux. To our knowledge, RSK representatives would be a very unnatural encoding for most tasks we consider. However, information about the longest increasing and decreasing subsequences of a permutation is easily accessible in this encoding.

There are many different ways of representing permutations, each of which is useful for different types of problems. In this paper, we use six different common representations of permutations. In the descriptions below, $\sigma: [n] \rightarrow [n]$ is a permutation in $S_n$.  
\begin{itemize}[leftmargin=1.5em, itemsep=0.25em]
    \item \textbf{One-line notation:} Here $\sigma$ is represented as the list $\sigma(1)\; \sigma(2) \; \dots \; \sigma(n)$, where the $i$th element in the list $j$ means that $\sigma(i) = j$. We have $\tau = 3 1 2 5 4$.
    \item \textbf{Cycle notation:} A permutation $\pi \in S_k$ is a \emph{cycle} if applying $\pi$ repeatedly to any $i \in [k]$ yields all elements of $[k]$. One way to write the \emph{cycle notation} for such a $\pi$ is then $(i, \pi(i),\dots, \pi^{k-1}(i))$. For example, the permutation $\pi = 2 3 1$ (in one-line notation) can be written as $\pi = (1,2,3)$ in cycle notation. Not all permutations are cycles but any permutation can be written as a disjoint product of cycles. For instance, $\tau = (1\;3\;2)(4\;5)$. There are many equivalent ways of ordering the disjoint cycles of a permutation, but the multiset of cycle sizes (known as the \emph{cycle type} of the permutation) is an invariant.
    \item \textbf{Inversion vector:} The \emph{inversion vector} representation of a permutation is an element $v_{\sigma} \in \mathbb{Z}_{\geq 0}^n$ where the $i$th element of $v_{\sigma}$ gives the number of elements larger than $i$ that appear before $i$ in one-line notation. The inversion vector for $\tau$ is $(1,1,0,1,0)$.
    \item \textbf{Lehmer code:} Similar to the inversion vector, this is a sequence of integers $(c_1,\dots,c_n)$ where $c_i$ is equal to the number of $j > i$ such that $\sigma(j) < \sigma(i)$. The Lehmer code for $\tau$ is $(2,0,0,1,0)$.
    \item \textbf{Reduced Coxeter expression:} Any permutation $\sigma$ can be written as a finite composition of Coxeter generators, $\sigma = s_{i_1}\dots s_{i_l}$. We call a product $s_{i_1}\dots s_{i_l}$ a \emph{reduced expression} of $\sigma$ when $l$ takes the minimum value among all such expressions. Reduced expressions are generally not unique. Both $s_4s_2s_1$ and $s_2s_4s_1$ are reduced expressions of $\tau$ since $s_2$ and $s_4$ commute.
    \item \textbf{RSK representative:} Applying the RSK algorithm to a permutation gives a unique representative of the permutation as a pair of standard Young tableaux. 
\end{itemize}

The \emph{strong Bruhat order} is a partial order, $\leq$, on the elements of $S_n$. For $\sigma, \pi \in S_n$, $\sigma \leq \pi$ if and only if some reduced word for $\sigma$ appears as a subword of a reduced word for $\pi$. The Bruhat order is intimately related to an important set of geometric constructions known as Schubert varieties \cite{fulton1997young} and was central to \cite{blundell2022towards}.

The study of \emph{permutation patterns}, which are relative orderings within a permutation, plays an important role in many applications. In this work, 9 different patterns appear in the training data, $213$-, $312$-, $132$-, $321$-, $1324$-, $1234$-, $2413$-, $4321$-, and $3412$-patterns. It is easiest to define what a pattern is using two of the specific instances above. $\sigma \in S_n$ contains a $321$-pattern if there are $i_1 < i_2 < i_3$ such that $\sigma(i_3) < \sigma(i_2) < \sigma(i_1)$. Similarly, $\sigma$ has a $3412$-pattern if there are $i_1 < i_2 < i_3 < i_4$ and $\sigma(i_3) < \sigma(i_4) < \sigma(i_1) < \sigma(i_2)$. Pattern avoidance connects permutations to other areas of math. $321$-avoiding permutations are counted by the Catalan numbers and are in bijection with a range of important constructions in combinatorics such as Dyck paths of length $2n$ \cite{stanley2015catalan}. $321$-avoiding permutations correspond to fully commutative elements in the symmetric group \cite{stembridge1996fully}. $3412$-avoidance is one part of a criterion for the smoothness of the corresponding Schubert variety \cite{lakshmibai1990criterion}.

The other tasks in PermuFormer's training set are listed in Tables \ref{table:property-volume} and \ref{table:operation-volume}, with background provided in Appendix \ref{sect:permutation-properties}.

\section{Training Data}
\label{sect:training_data}

To construct a training dataset for PermuFormer, we design a range of tasks related to permutations. These tasks fall into three broad families.
\begin{enumerate}[leftmargin=1.5em, itemsep=0.25em]
    \item {\textbf{Translation between permutation encodings:}} The model is provided with a permutation written in one encoding and is asked to produce the same permutation in another encoding. For example, given a permutation $\sigma$ written in one-line notation, write $\sigma$ as a reduced word. We sample from all 30 translation tasks that one can generate using the 6 different encodings above.
    \item {\textbf{Calculating permutation statistics:}} Given a permutation (in one of the 6 encodings), we ask the model to compute one of 32 statistics (Section \ref{sect:permutation-properties}). Since we can ask about these statistics for each of the 6 encodings, we get 192 different tasks for a fixed permutation.
    \item {\textbf{Performing basic operations or comparisons between permutations:}} Finally, we design 14 tasks that involve algebraic operations (such as composition, conjugation, etc.), combinatorial operations (like generating a permutation's complement), and comparison (such as determining whether one permutation is larger than another in Bruhat order). Taking into account encoding choice, this gives 84 different tasks for a single permutation (or pair of permutations). 
\end{enumerate}

We generate our training and evaluation datasets as follows. We sample uniformly from the three families of tasks (translation, statistic calculation, and operation/comparison). Once we have chosen one of these, we sample uniformly from the tasks within that task family. This includes randomly choosing an encoding type in the latter two cases. Finally, we sample permutations for the task. To make the data generation process less memory intensive, we only deduplicate within fixed 5-million-instance windows. Because there are far fewer permutations in $S_n$ when $n$ is small and because we sample uniformly from each symmetric group, most duplicates will likely involve small permutations (for instance, the total number of possible combinations of task/encoding/permutation triples for $S_8$ is over 19 billion, so extensive repeated sampling is unlikely). We believe this strategy is justified, since we want to balance the volume of large and small permutations.

The permutation groups we sample from include $S_2, \dots, S_{11}$ because (i) this is a wide enough range that PermuFormer learns representations of permutations across different sizes, (ii) many research-level problems can be explored in this regime, and (iii) symmetric groups on the larger side of the spectrum are large enough to avoid some degenerate behavior arising in small permutations.

Our training dataset contains 39.8 million instances with a total of $\sim2.66$ billion tokens (our tokenization scheme is discussed in Section \ref{sect-training}). The average number of tokens in an instance is $66.84$. This includes both the prompt and ground truth. We separately generate $200,001$ instances for evaluation. The training and test sets are strictly disjoint: no complete instance occurs in both train and test. We use SageMath for all permutation computations \cite{sagemath}. Data generation takes $\sim30$ minutes on a MacBook Pro. We provide detailed dataset statistics in Tables \ref{table:property-volume}, \ref{table:operation-volume}, and \ref{table:translation-volume}.

\subsection{Computation witnesses}

Early versions of PermuFormer struggled to predict some single-token properties such as the parity of a permutation. This is in contrast to other computationally demanding tasks like the RSK algorithm, where performance was better. One difference between these two tasks is that the computation required for the RSK algorithm is spread over many generated tokens. This observation is reminiscent of chain-of-thought and scratchpad strategies \cite{nye2021show,wei2022chain}, in which the model can use multiple tokens of generation to perform more complex computation. 

To aid in learning, we added computational witnesses to the expected output before final answer generation for some tasks. Thus, for the length task, we train the model to produce a reduced word for the permutation since the number of generators in this word is exactly the Coxeter length of the permutation. This resembles a simple form of chain-of-thought for these tasks. We list all tasks that use this computational witness approach and the nature of the witness in Table \ref{table:witnesses}.

\subsection{Data format}

PermuFormer is trained autoregressively like a language model. As such, we need to represent tasks as text. Because we are not interested in general linguistic ability, we opt for a formulaic approach to translating tasks into `sentences’. 

We introduce tokens that specify the beginning and end of each type of permutation encoding as well as different types of permutation properties and statistics. Thus, if we want to denote the permutation $\sigma = 3 1 2$ written in one-line notation, we write \verb|[one-line begin] 3 1 2 [one-line end]|. On the other hand, if we want to write $\sigma$ in cycle notation, we put \verb|[cycle notation begin] (1, 3, 2) [cycle notation end]|. Using tokens \verb|[encoding begin]| and \verb|[encoding end]| avoids ambiguity about the intended encoding. We use analogous delimiters when writing statistics. All instances begin with the size of the permutation (e.g. \verb|n7| for $\sigma \in S_7$) to eliminate ambiguity when translating from an encoding that does not implicitly specify the symmetric group that a permutation belongs to (e.g., from a Coxeter reduced expression to one-line notation).

Each instance is written as an equality. Because everything preceding the \verb|=| token will be treated as the prompt, the task needs to be fully specified on the left-hand side of the equality. Thus, for the translation task the target encoding \verb|[target encoding]| token precedes \verb|=|. We provide the templates used for the translation and property extraction tasks in Figure \ref{fig:example_color}, along with two examples. The format for operation-based tasks is equivalent.

\section{Tokenizer, Model, and Training}
\label{sect-training}

Since the training data for this model only includes a small set of words and symbols relative to natural language, we design a tokenization scheme with a vocabulary of 186 tokens. This includes integers, the delimiters discussed above, structural symbols like `,’ and `[‘, etc. PermuFormer is a 75-million-parameter LLaMA-style decoder-only transformer with 768-dimensional hidden representations, 12 blocks, 12 attention heads with group-query attention, an intermediate dimension size of 2,048, SwiGLU activations, and rotary positional embeddings with a theta value of 10,000. 

We trained PermuFormer autoregressively for 1,650,000 steps on an NVIDIA A100, with a batch size of 32, an initial learning rate of $1 \times 10^{-4}$, 100 warmup steps, and a weight decay value of 0.01. These parameters were chosen based on informal exploration and likely do not represent an optimal choice. We used a linear scheduler. Because our goal is for PermuFormer to learn about permutations (rather than generating permutations), we always start calculating the loss after the \verb|=| token, so that the model is only trained to generate the right side of the equation (see Figure \ref{fig:example_color} for an example).

\begin{figure}[ht]
\begin{center}
\begin{tcolorbox}[
    colback=gray!10,
    colframe=black,
    boxrule=0.8pt,
    arc=4pt,
    width=0.9\linewidth,
]
\small

\textbf{Translation template:}\\
\small{
{\ttfamily
[permutation size] [source encoding type begin] [permutation 1] [source encoding type end] [target encoding] =
[target encoding type begin] [permutation 2] [target encoding type end]
}}
\vspace{2mm}
\\
\textbf{Example:}\\
{\small
{\color{blue}\texttt{n5 1linebegin [1, 4, 5, 3, 2] 1lineend inversemake =}}
{\color{ForestGreen}\texttt{1linebegin [1, 5, 4, 2, 3] 1lineend}}
}
\vspace{0.5em}

\textbf{Property extraction template with witness}\\
\small{
{\ttfamily
[permutation size] [encoding type begin] [permutation] [encoding type end] [target statistic] =
[witnesses begin] [witnesses] [witnesses end] [statistic begin] [statistic] [statistic end]
}}
\vspace{2mm}
\\ 
\textbf{Example:}\\
{\small
{\color{blue}\texttt{n5 cyclenotationbegin [[1],[2,5,4,3]] cyclenotationend property avoidsmake [1,2,3,4] = }}{\color{ForestGreen}\texttt{witnessbegin [1,3,4,5] witnessend avoidsbegin False avoidsend}}
}

\end{tcolorbox}
\caption{Translation and property extraction templates and two examples. The prompts are in {\color{blue}{blue}} while the expected output is {\color{ForestGreen}{green}}}
\label{fig:example_color}
\end{center}
\end{figure}

\section{Performance}
\label{sect:performance}

\begin{figure}[htbp]
    \centering
    \includegraphics[width=0.85\textwidth]{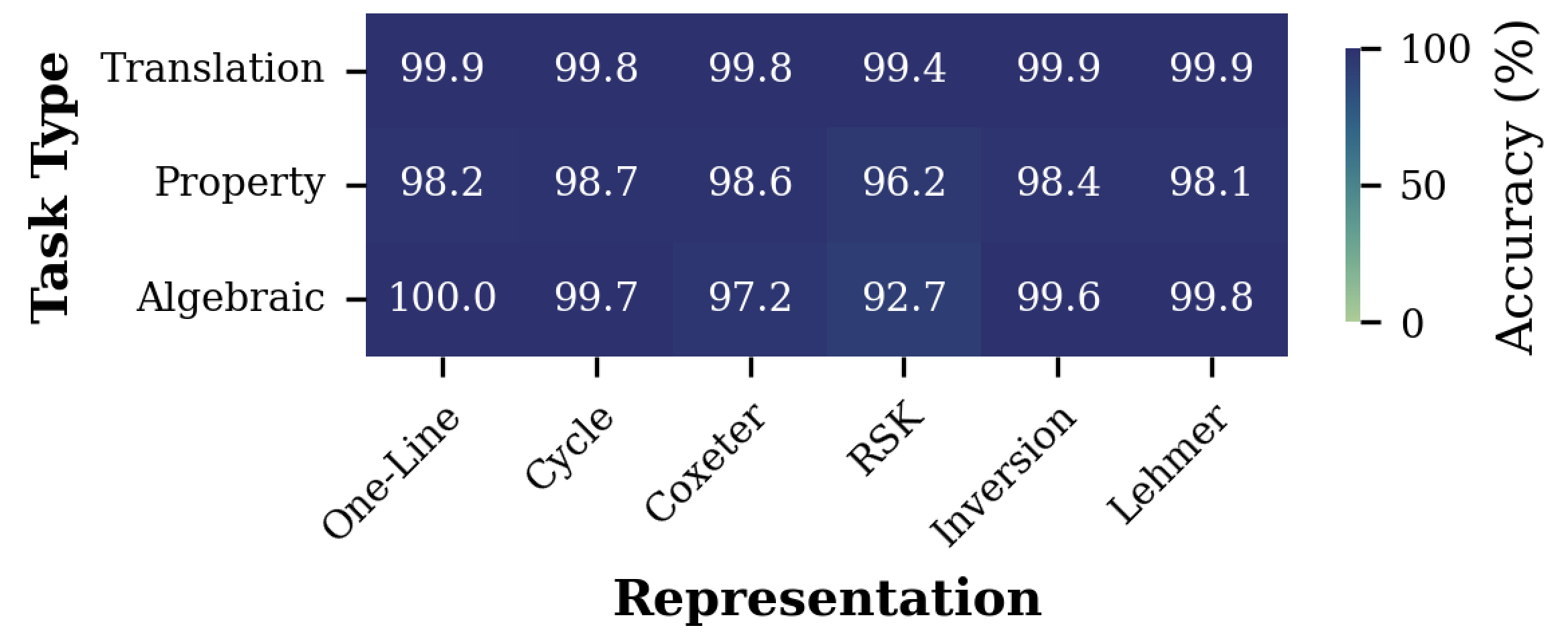}
    \caption{Average accuracy as a function of task family and encoding.}
    \label{fig:accuracy_across_tasks}
\end{figure}

We evaluate PermuFormer's performance based on exact match against ground truth. That is, we ask whether the content generated by PermuFormer exactly matches the ground-truth output token-for-token. Note that this requires specifying a canonical answer in cases where mathematically there may be many solutions. This is detailed in Section \ref{sect:multiple-correct-answers}. We call the fraction of instances where exact match holds for a task the task accuracy. Evaluated in this way, PermuFormer achieves high accuracy across the evaluation corpus. We visualize performance via a heatmap in Figure \ref{fig:accuracy_across_tasks}. A detailed breakdown of results can be found in Figures \ref{fig:accuracy_across_translation}, \ref{fig:accuracy_across_properties}, and \ref{fig:accuracy_across_operations} in the supplementary material.

\subsection{Analysis of results}

\textbf{Permutation encodings:} Across all task types, the RSK encoding proved to be the most challenging for PermuFormer. This was especially true on the algebraic tasks. This aligns with the practice of human mathematicians, where it would be very unnatural to use RSK representatives to perform all but a few of these tasks, especially tasks that involve permutation composition. It is interesting that this aspect of PermuFormer's performance mirrors human notions of difficulty. It is also worth noting that the Coxeter and RSK encodings are the least compact encodings in terms of token usage.

\begin{figure}[htbp]
    \centering
    \includegraphics[width=0.95\textwidth]{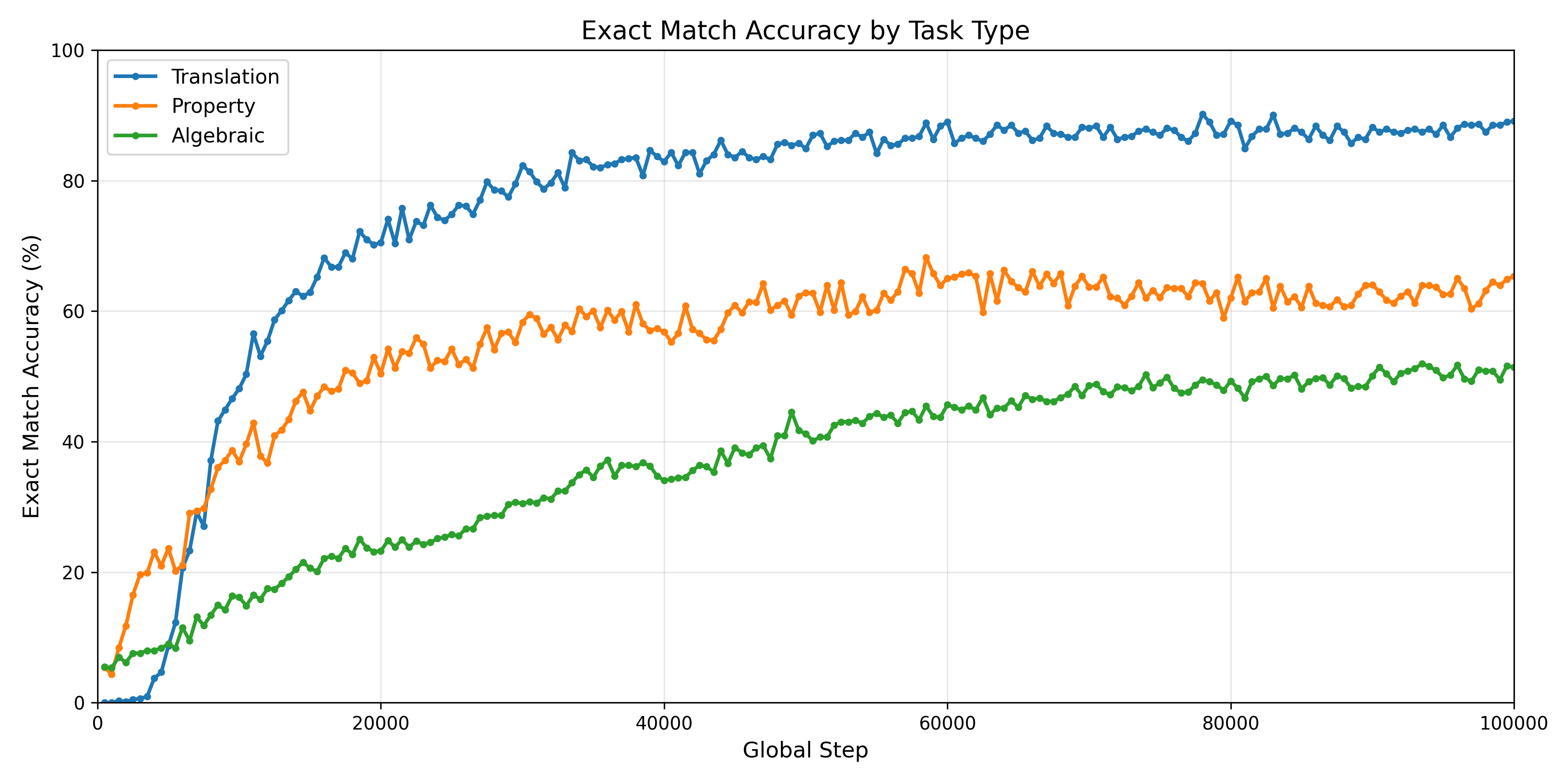}
    \caption{Average test accuracy over the first 100,000 training steps. Note that training continued out to 1,650,000 steps.}
    \label{fig:accuracy_over_training}
\end{figure}

\textbf{Translation:} PermuFormer learns translation tasks most quickly (Figure \ref{fig:accuracy_over_training}) and ultimately achieves a higher accuracy on them relative to the other task families (Figure \ref{fig:accuracy_across_tasks}), despite the fact that some of these tasks can be quite involved, for example, converting between RSK representatives and Coxeter reduced words. One might conjecture that learning proceeds quickly because the model finds an efficient parallelization strategy, allowing many independent, local computations, but the sequential nature of the RSK algorithm argues against this. Our current conjecture is that, because the answer requires many tokens, these types of problems offer a natural scratchpad for extended computation.

\textbf{Property extraction:} PermuFormer achieves high accuracy on most tasks, especially those that involve checking for a specific property (e.g., pattern avoidance or checking whether a permutation is a derangement). Tasks that are easily parallelized, such as identification of fixed points, are also quickly learned. Tasks that were more challenging include calculating length (unless the encoding is the reduced-word encoding), tasks involving longest increasing and decreasing subsequences (unless the encoding is the RSK representative), parity tasks, and cycle type (Figure \ref{fig:accuracy_across_properties}). These all involve tracking and consolidating a large amount of information, which may require sequential computation. They are also generally tasks where small errors can significantly change the answer. Learning improved substantially on these problems when witnesses were added.

\textbf{Operations:} As a family, algebraic operations took the longest for the model to learn (Figure \ref{fig:accuracy_over_training}). Those that were most challenging tended to require consecutive composition and inversion. The commutator ($\sigma^{-1}\tau^{-1}\sigma \tau$) and conjugation ($\tau \sigma \tau^{-1}$) tasks were learned slowly and the final model is still less reliable on these than it is on other tasks.

\subsection{Comparison to Frontier LLMs}

How hard are these tasks? Can they be solved by more capable generalist models? To test this, we randomly chose 150 instances from our evaluation set such that 50 involved permutations of size 5, 50 involved permutations of size 8, and 50 involved permutations of size 11. We then used Claude 4.6 Sonnet to create a document that explained how to interpret the syntax for these problems (see Section \ref{sect:llm_prompt}). We gave this, along with the left-hand side of each equality (e.g., the blue part of Figure \ref{fig:example_color}), to three near-frontier models \texttt{Claude 4.6 Sonnet}, \texttt{GPT-5.4}, and \texttt{Gemini 2.5 Pro} and asked them to complete the right-hand side without token limits. We did not penalize syntax errors made by the LLMs, programmatically extracting only the final answer.

As can be seen in Table \ref{table:frontier-performance}, recent Claude, GPT, and Gemini models only achieved moderate accuracy on these tasks. On the other hand, PermuFormer achieves perfect accuracy on those sampled problems involving mid-sized permutations. We caution that this is an imperfect comparison. The frontier language models that we tested had to operate in a unique syntax, which PermuFormer was specifically designed for. On the other hand, PermuFormer does not have access to test-time computation beyond that required for witness and answer generation, while the LLMs sometimes used a substantial number of tokens to answer a single question (e.g., $>$10,000 for Gemini). Thus, the primary point we hope to make with this experiment is that while these tasks may not be research-level (they can all be efficiently solved with SageMath or by a human with familiarity with algebraic combinatorics), they still require a nontrivial amount of computation. This makes them challenging for the LLMs tested here, which did not have access to tools.

\begin{table}[t]
\caption{Frontier model accuracy on 150 examples from PermuFormer's evaluation set.}
\label{table:frontier-performance}
\begin{center}
\begin{tabular}{lcccc}
 & \textbf{\small Claude 4.6 Sonnet} &
\textbf{\small GPT-5.4} & \textbf{\small Gemini 2.5 Pro} & \textbf{\small PermuFormer} \\
\midrule
  \textbf{$n=5$}   & 48\% & 32\% & 74\% & \textbf{100\%} \\
  \textbf{$n=8$}  & 52\% & 30\% & 62\% & \textbf{100\%} \\
  \textbf{$n=11$}  & 42\% & 20\% & 54\% & \textbf{92\%} \\

\end{tabular}
\end{center}
\end{table}

\subsection{Fine-tuning PermuFormer for Tasks Outside of the Training Set}
\label{sect:fine-tuning}

One of the primary motivations for developing PermuFormer is to provide the community with a small base transformer that captures rich representations of permutations and can be easily fine-tuned for specific research tasks. We describe experiments that demonstrate this in this section. We test this using both tasks unseen during pretraining (new pattern avoidance, mHeight) and research-level problems (Kazhdan-Lusztig polynomial coefficients and degree and Schubert polynomial structure constants). The Schubert polynomial structure constant, mHeight, and Kazhdan-Lusztig tasks were inspired by \cite{chau2025machine}, though the exact data differs slightly to take into account our tokenization scheme. We aim to show that the combinatorics-specific pretraining we perform helps a model generalize to other combinatorics tasks that involve permutations. We therefore compare PermuFormer's performance against four baselines: (i) a randomly initialized model with the same architecture as PermuFormer, (ii) EleutherAI's \texttt{pythia-70m-deduped}, a generic pretrained language model of comparable size \cite{biderman2023pythia}, (iii) a basic MLP, and (iv) a logistic regression model. 

The comparison to (i) is designed to study whether pretraining actually offers benefits over simply training from scratch on a target dataset while controlling for architecture. We compare to (ii) to probe whether it is specifically exposure to permutation-focused tasks that is important or whether general language training is sufficient. \texttt{pythia-70m-deduped} has 70 million parameters compared to PermuFormer’s 75 million. In our experiments for (iii), we use a basic ReLU MLP architecture with three hidden layers and a hidden dimension of 256. Since the MLP cannot accept a token sequence, we remove all nonessential structural, delimiter, and property-indicator tokens. We then one-hot-encode each remaining token. We concatenate these and pad the resulting vector so that each instance has the same size. This comparison is designed to test PermuFormer’s performance against a model which is not a transformer. \cite{chau2025machine}, for instance, found that MLPs were sometimes significantly easier than transformers to train. 
Finally, one might ask whether PermuFormer is a better base to fine-tune from only because it has access to the statistics it learned during pretraining. To explore this, we vectorized each dataset as we did when training the MLP, but we additionally concatenated a vector recording each property found in the pretraining set for each of the permutations appearing in the instance. This includes numeric features that were encoded via their numeric value, Boolean features that were encoded as a 0 or 1, and lists that were encoded by a sequence of numeric features which were padded to ensure uniform length.  We then trained a logistic regression model on the resulting dataset. If PermuFormer uses properties learned during pretraining to solve tasks in a linear manner, then we expect strong performance from the logistic regression model.

We describe the fine-tuning tasks below. For each of these we made choices about how to represent input for the task within PermuFormer's vocabulary. In all cases we re-used existing tokens rather than adding new ones. To isolate the effect of pretraining, we also chose to provide the exact same input format to all transformers. We ran separate experiments using each of the six different permutation encodings. We chose dataset sizes based on the difficulty of the problem. For problems that are relatively easy such as pattern avoidance, we used small datasets. For harder, research-level problems we used larger datasets. Further details about the fine-tuning set-up for each task can be found in Section \ref{sect:finetuning_hyperparameters}.

\textbf{Unseen pattern avoidance:} PermuFormer is pretrained to determine whether permutations avoid the patterns $213$, $312$, $132$, $321$, $1324$, $1234$, $2413$, $4321$, $2143$, and $3412$. The first two fine-tuning tasks focus on detecting whether a permutation avoids $231$ and $1342$, respectively. We generate small, class-balanced datasets with permutations from $S_9$. The training set for $231$-patterns contains 241 elements while the training set for $1342$-patterns contains 477 elements. 

\textbf{mHeight:} mHeight is a permutation statistic derived from $3412$-patterns. It played a fundamental role in a proof \cite{gaetz2024minimal} that resolved a longstanding conjecture of Billey and Postnikov \cite{billey2005smoothness}. To calculate it, one finds all instances $(w_p,w_q,w_r,w_s)$ of the pattern $3412$ in a permutation $\sigma$ and then finds the maximum value of $w_p - w_s$. Note that, while PermuFormer is trained to detect whether a permutation has a $3412$-pattern, calculating mHeight requires finding all such instances, calculating $w_p - w_s$ for each, and finding the maximum of these values. We construct a training dataset with 12,159 examples balanced between no $3412$-pattern, mHeight of 1, and mHeight of 2. We use permutations from $S_9$.

\textbf{Schubert polynomial structure constants:} Let $\sigma$ be a permutation. The Schubert polynomial $\mathcal{S}_\sigma$ is a polynomial in finitely many variables $x_1,x_2,\dots$. Schubert polynomials hold deep combinatorial and geometric significance \cite{manivel2001symmetric}. They carry, for example, structural information about the geometry of flag manifolds.

Despite intense study, the structure constants $c^\pi_{\sigma,\tau}$, which arise in the expansion
\begin{equation*}
    \mathcal{S}_\sigma \mathcal{S}_\tau = \sum_{\pi \in S_{\infty}} c^{\pi}_{\sigma,\tau}\mathcal{S}_\pi,
\end{equation*}
where $S_\infty = \bigcup_n S_n$ under the standard embeddings, remain poorly understood from a combinatorial perspective. Predicting $c^\pi_{\sigma,\tau}$ (which is always a non-negative integer) given $\sigma$, $\tau$, and $\pi$ is thus a demanding task for which there is no known closed-form solution. Since most $c^\pi_{\sigma,\tau}$ are zero, we use the sampling approach from \cite{chau2025machine} to create a dataset with 103,139 triples of permutations along with their associated structure constants. This breaks down as 39,342 instances with value 0, 39,342 instances with value 1, 23,938 instances with value 2, and 517 instances with value 3 or greater. Sampling details can be found in Section \ref{sect:finetuning_hyperparameters}.

\textbf{Kazhdan-Lusztig polynomials:} Kazhdan-Lusztig polynomials are a family of polynomials $P_{\sigma,\tau}(q)$ in the variable $q$ that are indexed by two permutations $\sigma$ and $\tau$ belonging to the same symmetric group \cite{kazhdan1979representations}. They carry significant geometric and representation-theoretic information, yet despite over 40 years of study many of their properties remain mysterious. Calculating the degree of $P_{\sigma,\tau}(q)$ using $\sigma$ and $\tau$ is complex, with no closed-form solution. 

We design two different fine-tuning tasks around Kazhdan-Lusztig polynomials: predicting the degree of the polynomial and predicting its coefficients. We use code from \cite{warringtonwebsite} to generate 50,000 Kazhdan-Lusztig polynomials corresponding to permutations from $S_9$ and then subsample for each of the tasks. For instance, for the degree prediction task we downselect to obtain 7,589 examples that are zero, 7,589 that are constant, 7,589 that have degree 1, 7,589 that have degree 2, and 4,853 that have degree 3 or greater. We then provide models with $\sigma$ and $\tau$ and have them predict $\deg(P_{\sigma,\tau}(q))$.

We follow a similar procedure for our coefficient prediction task. For a fixed $k$, the task is to predict the coefficient on $q^k$ in $P_{\sigma,\tau}(q)$ given $\sigma$ and $\tau$. Combinatorial interpretations for many of these coefficients are unknown. We run experiments for integers $0 \leq k \leq 4$. For $k > 4$, the number of non-zero examples becomes very small.

\begin{table}[t]
\caption{\small{Accuracy after fine-tuning PermuFormer on tasks not seen during training (rows) in terms of permutation encoding. The best score(s) for each setting are in bold. We include 95\% confidence intervals based on three independent fine-tuning runs.}}
\label{table:rep-instance}
\begin{center}
\setlength{\tabcolsep}{3pt}
\scriptsize
\begin{tabular}{llcccccc}
\textbf{Task} & \textbf{Model} &
\textbf{one-line} & \textbf{Cycle} & \textbf{Inversion} & \textbf{Reduced} & \textbf{RSK} & \textbf{Lehmer} \\
\midrule

\multicolumn{8}{c}{\textbf{Equivalent to training task}} \\ \cmidrule(lr){1-8}

\multirow{5}{*}{KL coeff. 0}
  & PermuFormer & \textbf{99.9$\pm$0.1\%} & \textbf{99.6$\pm$0.1\%} & \textbf{99.8$\pm$0.6\%} & \textbf{99.6$\pm$0.5\%} & \textbf{99.4$\pm$0.4\%} & \textbf{99.8$\pm$0.5\%} \\
  & From scratch & 98.5$\pm$1.4\% & 94.2$\pm$1.7\% & 97.4$\pm$0.6\% & 90.9$\pm$0.9\% & 85.8$\pm$2.8\% & 97.6$\pm$0.8\% \\
  & Pythia-70m & 95.2$\pm$7.8\% & 94.5$\pm$1.7\% & 94.3$\pm$8.8\% & 90.4$\pm$3.8\% & 86.3$\pm$7.4\% & 93.2$\pm$1.4\% \\
  & MLP & 98.0$\pm$0.4\% & 86.9$\pm$0.6\% & 97.2$\pm$0.4\% & 91.4$\pm$0.1\% & 87.3$\pm$0.9\% & 96.0$\pm$1.5\% \\
  & Logistic regression & 90.6\% & 90.4\% & 90.4\% & 90.2\% & 90.6\% & 90.6\% \\
\midrule

\multicolumn{8}{c}{\textbf{Related to pretraining tasks}} \\ \cmidrule(lr){1-8}

\multirow{5}{*}{Avoids 231}
  & PermuFormer & 94.9$\pm$0.5\% & 90.9$\pm$1.1\% & 96.3$\pm$0.0\% & 94.0$\pm$1.9\% & \textbf{98.0$\pm$0.5\%} & 94.3$\pm$0.5\% \\
  & From scratch & 94.4$\pm$2.5\% & 83.4$\pm$9.7\% & 94.8$\pm$6.1\% & 90.1$\pm$5.1\% & 85.3$\pm$1.4\% & 86.3$\pm$2.8\% \\
  & Pythia-70m & 92.3$\pm$7.6\% & 84.5$\pm$4.7\% & 90.5$\pm$3.0\% & 84.0$\pm$1.1\% & 81.9$\pm$4.7\% & 85.4$\pm$1.9\% \\
  & MLP & 53.6$\pm$0.0\% & 47.9$\pm$0.0\% & 47.6$\pm$0.0\% & 49.4$\pm$0.0\% & 48.7$\pm$0.0\% & 49.1$\pm$0.0\% \\
  & Logistic regression & \textbf{95.5\%} & \textbf{94.0\%} & \textbf{96.3\%} & \textbf{95.9\%} & 91.8\% & \textbf{97.8\%} \\
\midrule

\multirow{5}{*}{Avoids 1342}
  & PermuFormer & 86.4$\pm$1.4\% & \textbf{77.7$\pm$1.4\%} & \textbf{89.8$\pm$0.8\%} & \textbf{88.2$\pm$0.7\%} & \textbf{88.2$\pm$0.7\%} & \textbf{90.1$\pm$0.5\%} \\
  & From scratch & \textbf{86.6$\pm$3.3\%} & 68.6$\pm$2.4\% & 80.9$\pm$3.9\% & 76.5$\pm$1.2\% & 72.1$\pm$1.5\% & 85.0$\pm$1.0\% \\
  & Pythia-70m & 74.7$\pm$4.6\% & 68.9$\pm$10.5\% & 71.1$\pm$18.4\% & 73.8$\pm$13.6\% & 67.5$\pm$2.3\% & 79.8$\pm$6.8\% \\
  & MLP & 50.0$\pm$0.0\% & 50.0$\pm$0.0\% & 50.0$\pm$0.0\% & 50.0$\pm$0.0\% & 50.0$\pm$0.0\% & 50.0$\pm$0.0\% \\
  & Logistic regression & 75.1\% & 68.3\% & 76.8\% & 72.3\% & 64.7\% & 72.1\% \\
\midrule


\multirow{5}{*}{mHeight}
  & PermuFormer & 95.4$\pm$1.1\% & \textbf{85.2$\pm$4.1\%} & \textbf{96.4$\pm$0.6\%} & \textbf{92.7$\pm$0.8\%} & \textbf{92.2$\pm$1.2\%} & \textbf{94.9$\pm$0.4\%} \\
  & From scratch & \textbf{98.9$\pm$1.3\%} & 55.2$\pm$3.1\% & 91.8$\pm$3.0\% & 73.3$\pm$4.9\% & 64.1$\pm$2.8\% & 92.3$\pm$3.0\% \\
  & Pythia-70m & 85.9$\pm$10.0\% & 59.3$\pm$2.0\% & 82.9$\pm$2.3\% & 66.1$\pm$32.1\% & 70.9$\pm$4.1\% & 83.2$\pm$1.7\% \\
  & MLP & 66.5$\pm$3.5\% & 49.8$\pm$2.1\% & 70.7$\pm$5.2\% & 53.2$\pm$9.2\% & 57.8$\pm$1.3\% & 66.2$\pm$24.1\% \\
  & Logistic regression & 85.9\% & 83.7\% & 84.6\% & 83.0\% & 82.1\% & 86.6\% \\
\midrule

\multicolumn{8}{c}{\textbf{Research-level tasks}} \\ \cmidrule(lr){1-8}

\multirow{5}{*}{Schubert}
  & PermuFormer & \textbf{92.3$\pm$4.9\%} & \textbf{84.4$\pm$7.5\%} & \textbf{93.5$\pm$4.6\%} & \textbf{92.4$\pm$1.2\%} & \textbf{75.4$\pm$2.5\%} & 79.7$\pm$34.6\% \\
  & From scratch & 65.4$\pm$3.8\% & 79.1$\pm$2.1\% & 66.3$\pm$4.2\% & 82.9$\pm$11.0\% & 61.1$\pm$2.2\% & 62.9$\pm$1.3\% \\
  & Pythia-70m & 60.8$\pm$1.2\% & 61.9$\pm$5.4\% & 66.0$\pm$8.8\% & 84.9$\pm$2.9\% & 61.3$\pm$0.3\% & 62.8$\pm$2.4\% \\
  & MLP & 86.4$\pm$28.0\% & 79.3$\pm$3.5\% & 83.8$\pm$31.0\% & 85.4$\pm$0.1\% & 66.5$\pm$0.9\% & \textbf{90.9$\pm$2.3\%} \\
  & Logistic regression & 68.4\% & 67.4\% & 67.9\% & 67.3\% & 68.2\% & 68.0\% \\
\midrule

\multirow{5}{*}{KL degree}
  & PermuFormer & 65.1$\pm$5.3\% & 56.7$\pm$1.6\% & \textbf{65.6$\pm$1.8\%} & \textbf{60.7$\pm$2.3\%} & \textbf{59.8$\pm$3.6\%} & \textbf{65.2$\pm$2.9\%} \\
  & From scratch & 56.8$\pm$8.9\% & 36.3$\pm$1.4\% & 48.3$\pm$6.1\% & 35.0$\pm$3.9\% & 29.3$\pm$11.5\% & 47.6$\pm$4.6\% \\
  & Pythia-70m & \textbf{73.4$\pm$5.3\%} & 55.6$\pm$5.4\% & 60.3$\pm$11.0\% & 52.1$\pm$1.7\% & 46.1$\pm$15.9\% & 58.8$\pm$2.0\% \\
  & MLP & 66.7$\pm$1.2\% & 46.1$\pm$1.4\% & 65.1$\pm$1.7\% & 52.5$\pm$0.9\% & 46.8$\pm$2.4\% & 63.9$\pm$1.7\% \\
  & Logistic regression & 59.0\% & \textbf{57.4\%} & 56.8\% & 58.6\% & 57.6\% & 59.3\% \\
\midrule

\multirow{5}{*}{KL coeff. 1}
  & PermuFormer & \textbf{89.5$\pm$3.4\%} & \textbf{76.9$\pm$3.4\%} & \textbf{89.3$\pm$2.7\%} & \textbf{79.5$\pm$2.1\%} & \textbf{80.8$\pm$1.6\%} & \textbf{87.6$\pm$4.5\%} \\
  & From scratch & 77.7$\pm$1.6\% & 61.2$\pm$0.7\% & 69.8$\pm$1.1\% & 60.6$\pm$0.5\% & 58.1$\pm$5.9\% & 68.3$\pm$10.8\% \\
  & Pythia-70m & 69.1$\pm$4.1\% & 60.6$\pm$4.0\% & 61.0$\pm$2.3\% & 60.6$\pm$0.9\% & 59.5$\pm$2.7\% & 61.7$\pm$6.9\% \\
  & MLP & 69.7$\pm$1.7\% & 56.2$\pm$0.8\% & 66.9$\pm$1.7\% & 60.6$\pm$1.2\% & 57.6$\pm$1.6\% & 68.0$\pm$1.2\% \\
  & Logistic regression & 65.3\% & 64.3\% & 65.3\% & 66.2\% & 65.5\% & 65.0\% \\
\midrule

\multirow{5}{*}{KL coeff. 2}
  & PermuFormer & \textbf{67.2$\pm$1.0\%} & \textbf{64.4$\pm$1.1\%} & \textbf{70.1$\pm$2.6\%} & \textbf{64.2$\pm$0.3\%} & \textbf{67.6$\pm$1.3\%} & \textbf{69.0$\pm$2.1\%} \\
  & From scratch & 63.5$\pm$1.0\% & 62.5$\pm$0.9\% & 64.0$\pm$2.5\% & 62.0$\pm$0.2\% & 58.2$\pm$2.2\% & 62.7$\pm$2.0\% \\
  & Pythia-70m & 62.1$\pm$0.9\% & 62.8$\pm$0.7\% & 63.2$\pm$1.4\% & 61.8$\pm$0.3\% & 55.8$\pm$7.8\% & 61.9$\pm$3.2\% \\
  & MLP & 61.7$\pm$0.8\% & 56.4$\pm$0.5\% & 64.1$\pm$1.9\% & 61.7$\pm$0.3\% & 55.6$\pm$0.8\% & 62.9$\pm$0.4\% \\
  & Logistic regression & 61.4\% & 60.2\% & 61.2\% & 60.2\% & 61.2\% & 61.2\% \\
\midrule

\multirow{5}{*}{KL coeff. 3}
  & PermuFormer & \textbf{70.0$\pm$2.1\%} & \textbf{65.5$\pm$0.4\%} & \textbf{68.5$\pm$1.2\%} & \textbf{67.4$\pm$0.3\%} & \textbf{66.9$\pm$3.6\%} & \textbf{69.7$\pm$6.4\%} \\
  & From scratch & 63.8$\pm$0.9\% & 63.6$\pm$1.2\% & 63.0$\pm$1.5\% & 63.5$\pm$0.2\% & 58.6$\pm$3.0\% & 63.2$\pm$0.2\% \\
  & Pythia-70m & 63.4$\pm$0.2\% & 63.7$\pm$1.5\% & 62.6$\pm$0.9\% & 63.6$\pm$0.2\% & 60.1$\pm$0.2\% & 63.8$\pm$1.3\% \\
  & MLP & 64.0$\pm$1.7\% & 61.8$\pm$1.0\% & 63.4$\pm$1.5\% & 63.5$\pm$0.4\% & 59.0$\pm$0.7\% & 65.2$\pm$1.0\% \\
  & Logistic regression & 63.6\% & 63.3\% & 63.9\% & 62.9\% & 63.9\% & 63.8\% \\
\midrule

\multirow{5}{*}{KL coeff. 4}
  & PermuFormer & \textbf{65.0$\pm$1.1\%} & \textbf{64.5$\pm$2.0\%} & \textbf{68.1$\pm$0.8\%} & \textbf{64.9$\pm$1.1\%} & \textbf{66.8$\pm$2.0\%} & \textbf{66.2$\pm$1.0\%} \\
  & From scratch & 63.2$\pm$0.6\% & 63.1$\pm$0.8\% & 66.2$\pm$0.8\% & 64.3$\pm$0.9\% & 61.8$\pm$1.5\% & 64.4$\pm$0.6\% \\
  & Pythia-70m & 63.6$\pm$0.5\% & 64.1$\pm$0.7\% & 66.3$\pm$0.8\% & 64.3$\pm$0.0\% & 63.2$\pm$0.9\% & 64.6$\pm$0.3\% \\
  & MLP & 63.9$\pm$1.1\% & 59.2$\pm$1.1\% & 66.1$\pm$0.7\% & 64.0$\pm$0.3\% & 60.0$\pm$0.6\% & 64.7$\pm$0.4\% \\
  & Logistic regression & 62.0\% & 60.9\% & 62.4\% & 63.7\% & 62.4\% & 61.6\% \\
\midrule

\end{tabular}
\end{center}
\end{table}

\subsection{Analysis of Fine-tuning Results} The results of our fine-tuning experiments are shown in Table \ref{table:rep-instance}. Since all tasks are framed as classification problems, we measure performance based on accuracy. Unlike our in-distribution evaluation procedure, where we judge correctness based on the entire generated token sequence, for fine-tuning tasks we only extract the numerical or Boolean answer. This allows us to avoid penalizing models for improper syntax. We provide 95\% confidence intervals based on three independent fine-tuning runs. For pretrained PermuFormer and \texttt{pythia-70m-deduped}, runs differed in batch ordering. For PermuFormer trained from scratch and MLPs, we also randomly initialized weights for each run. We do not report confidence intervals for the logistic regression model since it is deterministic.

We discuss each fine-tuning task and note some trends below. 

\textbf{Kazhdan-Lusztig coefficient on $q^0$:} The constant term in a Kazhdan-Lusztig polynomial is 1 if $\sigma \leq \tau$ and otherwise it is zero. Thus, this fine-tuning task is equivalent to predicting whether $\sigma \leq \tau$, a task that appears in the pretraining set. It is therefore not surprising that PermuFormer achieves near-perfect accuracy across encodings. 

\textbf{Avoids 231:} Here our logistic regression model performs best on all but the RSK encoding, where PermuFormer achieves 98.0\% accuracy. We conjecture that for the simple task of detecting 231-patterns, the collection of statistics that the logistic regression model has direct access to contains strong correlates that may give it high sample efficiency. Recall that this is a very small training set with only 241 examples. We found that when we increased the training set size, all models were able to reach near-perfect accuracy. Among models without direct access to statistics, PermuFormer achieved the best score across all encodings. 

\textbf{Avoids 1342:} This task involves a more complex pattern. Pretrained PermuFormer outperforms all other models across encodings with the exception of one-line notation, where PermuFormer trained from scratch exceeds pretrained PermuFormer by 0.2\%. The logistic regression model, which achieved the highest performance on almost all encodings for the task of detecting $231$-patterns, here performs worse than the transformers, suggesting that linear prediction from a large bank of features is insufficient for strong performance on some tasks. This trend continues in many of the tasks below.

\textbf{mHeight:} The mHeight task is rooted in pattern detection, specifically the ability to identify $3412$-patterns. However, rather than just asking whether a $3412$-pattern exists, it asks for a maximal statistic over all occurrences. Pretrained PermuFormer performs better than other baselines on all but one-line notation, where PermuFormer trained from scratch performs better (98.9\% versus 95.4\%). Notably, PermuFormer trained from scratch drops to 55.2\% when we move to cycle notation while pretrained PermuFormer maintains 85.2\% accuracy, reinforcing PermuFormer’s robustness across permutation encodings.

\textbf{Schubert polynomial structure constants:} On this task, pretrained PermuFormer achieves the highest accuracy across all encodings except for Lehmer code, where the MLP scored better. This is a task where pretrained PermuFormer substantially outperforms all other transformers across encodings (e.g., on one-line notation pretrained PermuFormer achieves 92.3\% accuracy while PermuFormer trained from scratch achieves 65.4\% accuracy and fine-tuned Pythia achieves 60.8\% accuracy). 

\textbf{Kazhdan-Lusztig polynomial degree:} Pretrained PermuFormer achieves the highest score on four of the six encodings. The Pythia model achieves the highest accuracy on one-line notation encoding while logistic regression scores highest on cycle notation. Overall, scores are low on this task. It may be that this is a fundamentally difficult task that requires more data, larger models, or more varied pretraining (in the case of PermuFormer).

\textbf{Kazhdan-Lusztig coefficients on $q^1$, $q^2$, $q^3$, and $q^4$:} Pretrained PermuFormer achieves the highest accuracy across all models when predicting the coefficients on $q^1$, $q^2$, $q^3$, and $q^4$. The amount by which pretrained PermuFormer exceeds the other models in accuracy tends to decrease as we increase the degree of the coefficient. We conjecture that higher-degree coefficients may require features that go beyond PermuFormer’s pretraining set. It is also possible that they would benefit from a larger model.

In addition to achieving the highest score on the majority of tasks above, pretrained PermuFormer is also relatively robust across encodings. This is a valuable feature in research mathematics, since we often do not know the most useful encoding ahead of time. 

\section{How Does PermuFormer Perform Computation?}

In this section, we use standard tools from neural network interpretability to better understand how PermuFormer performs tasks.

\subsection{Some statistics are captured in the prompt representation, others are extracted during generation} 
\label{sect:statistics-built-into-prompt}
PermuFormer is capable of extracting a range of different statistics from permutations. Are these immediately available to the model after processing the prompt, or does the model require computation through a sequence of generated tokens? To explore this question, we use linear probing to identify at what point an answer can be linearly decoded from a model's activations. To make this process simpler, we probe tasks where the answer is a single token. This includes tasks with a Boolean answer (e.g., predicting pattern avoidance) and tasks with an integer answer (e.g., predicting Coxeter length).

\begin{figure}[htbp]
    \centering
    \includegraphics[width=0.8\textwidth]{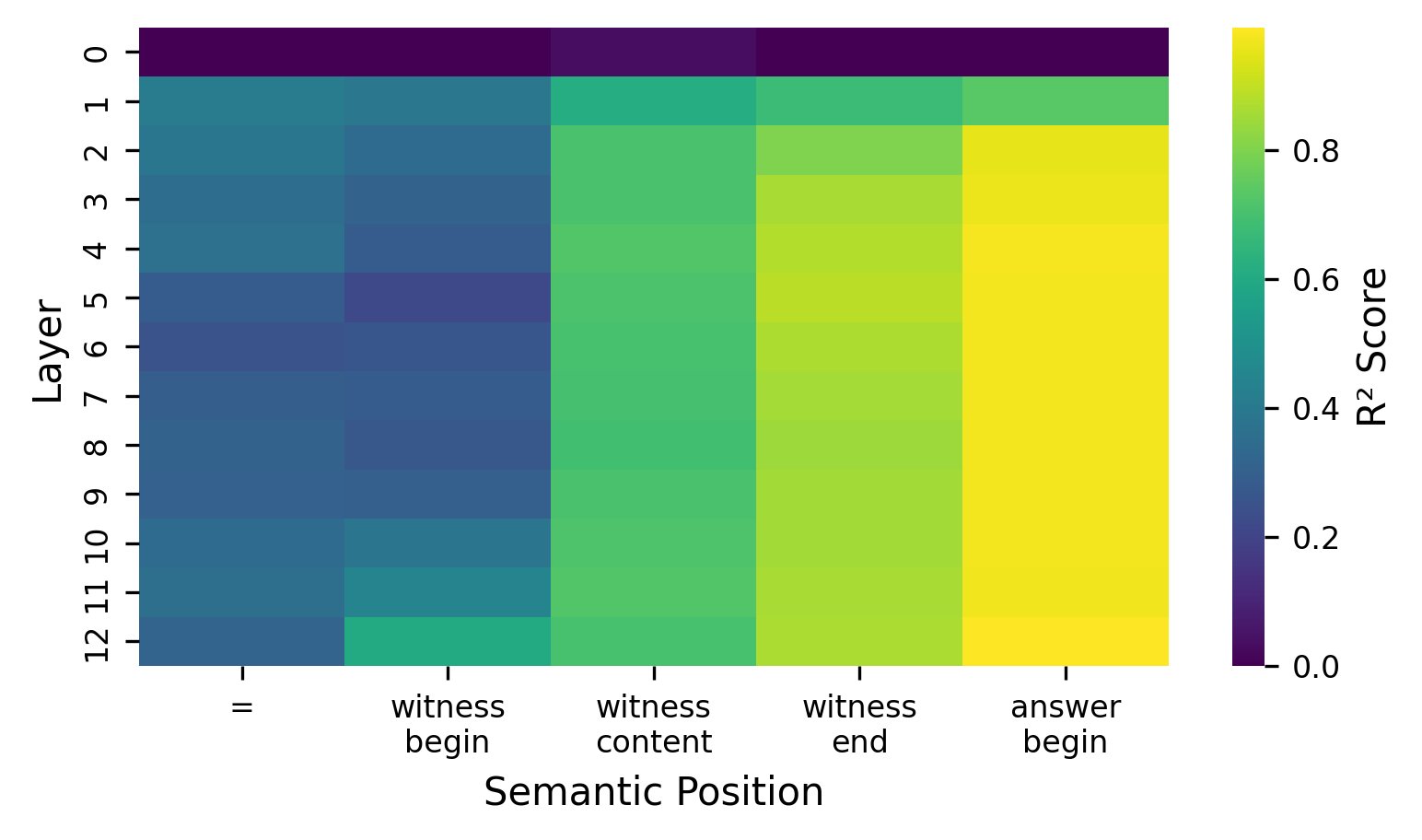}
    \includegraphics[width=0.8\textwidth]{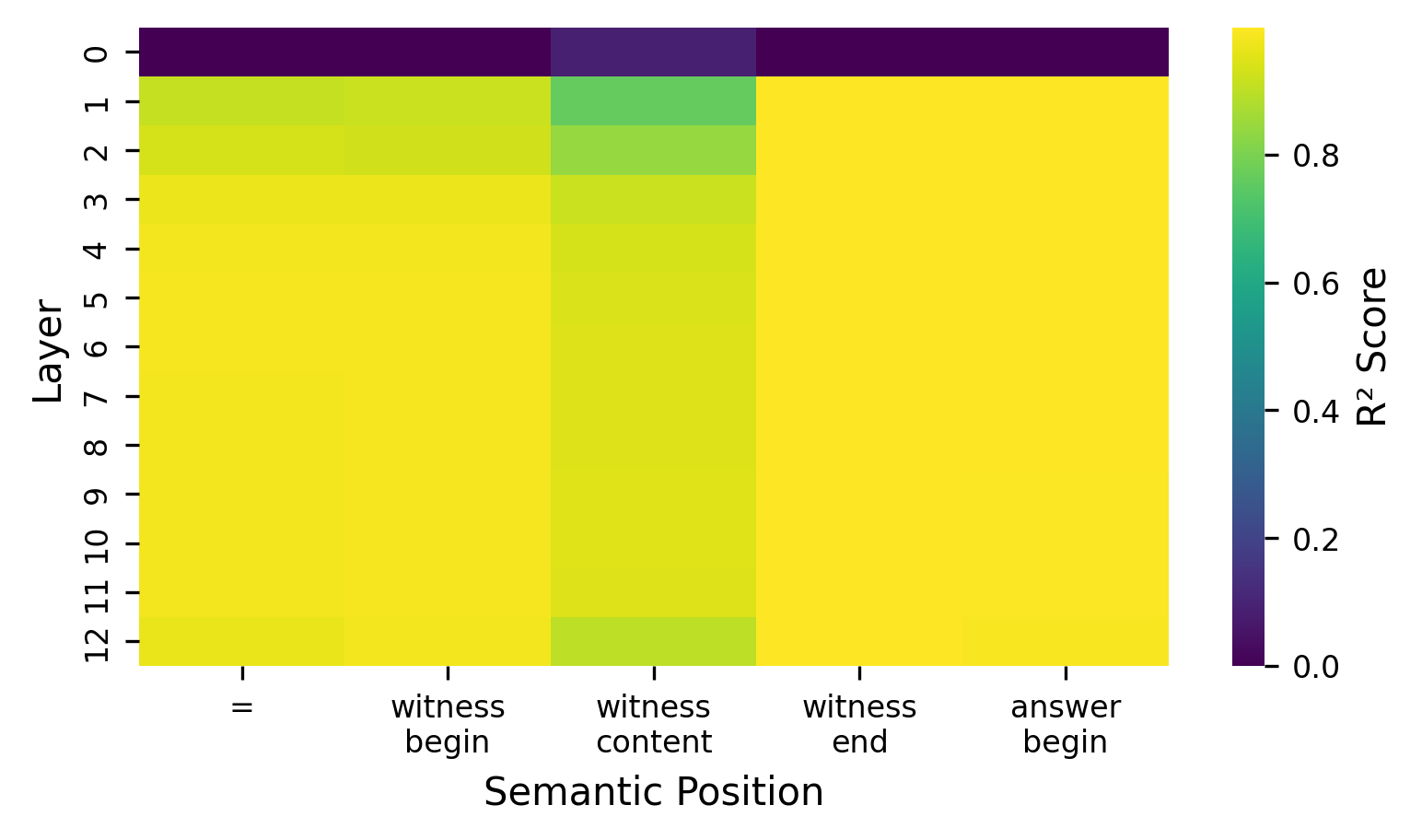}
    \includegraphics[width=0.8\textwidth]{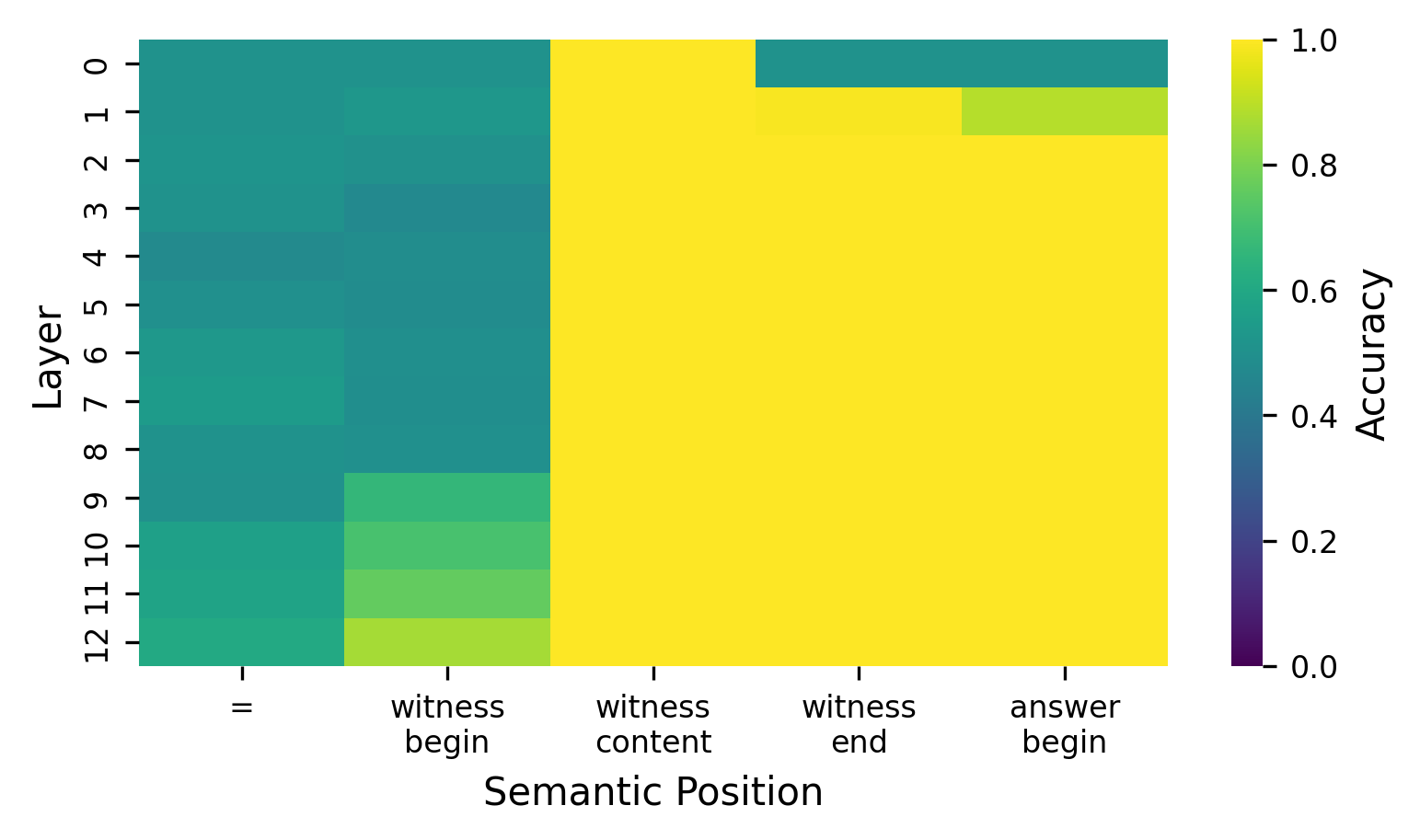}
    \caption{Heatmaps representing test $R^2$ scores (for regression) and accuracy values (for classification) for probes trained on specific layers (rows) and token positions (columns). \textbf{(Top)} Predicting the order of a permutation. \textbf{(Middle)} Predicting the Coxeter length of a permutation. \textbf{(Bottom)} Predicting the sign of a permutation.}
    \label{fig:heatmap-order}
\end{figure}

Because of the way that we designed PermuFormer's output format, it does not generate the answer directly from the prompt. Instead, generation is broken down into five groups of tokens. These five groups are consistent across all the tasks that we consider in this analysis. Specifically, we probe the following tokens or groups of tokens: (1) The \texttt{=} token, (2) the \texttt{[witnessbegin]} token, (3) an average across witness tokens, (4) the \texttt{[witnessend]} token, and (5) the beginning delimiter token for the answer (e.g., \texttt{avoidsbegin}). The \texttt{=} token is the final token in the prompt, so this is the model's representation of the full prompt prior to generating the first token. On the other hand, the beginning delimiter token for the answer is the final representation before the answer token itself is generated. If the model can effectively solve a task, we expect probing to achieve high accuracy here.

Our analysis shows that patterns in probing performance differ substantially from task to task. For some tasks, probing achieves high scores early in generation, suggesting that the solution to the task is accessible in the model's initial representation of the input. In other cases, probes only detect the answer during or after the witness tokens have been generated. Three examples are shown in Figure \ref{fig:heatmap-order}. The top heatmap corresponds to calculating the order of a permutation, the middle heatmap corresponds to calculating the length of a permutation, and the bottom heatmap corresponds to calculating the parity of a permutation. Each cell is the average $R^2$ or accuracy score on a 435, 414, and 436 instance test set for order, length, and parity respectively. A unique probe is trained per layer and per token position with training set sizes of 1,738, 1,654, and 1,743 instances respectively. When calculating order, we see that probe performance only gradually improves over generation. Even at the end of witness generation, $R^2$ scores hover around 80\%. On the other hand, length is linearly decodable before the first token has even been generated.

\subsection{Closely related tasks can emerge at different stages of generation} Comparing the calculation of length and parity (middle and bottom heatmaps in Figure \ref{fig:heatmap-order}) is interesting since length and parity are closely related. Suppose that 
\begin{equation*}
    \operatorname{Inv}(\sigma) = \{(i,j) \;|\; \sigma(j) < \sigma(i),\;1 \leq i < j \leq n\}
\end{equation*}
is the full set of inversions in a permutation $\sigma \in S_n$. Then length is the size of $\operatorname{Inv}(\sigma)$, $|\operatorname{Inv}(\sigma)|$, while parity is even if $|\operatorname{Inv}(\sigma)|$ is even and odd if $|\operatorname{Inv}(\sigma)|$ is odd. So, parity is a simple function of length, but length is linearly decodable immediately from the initial representation of the prompt while parity only becomes linearly decodable when witness tokens are generated. Given how closely they are related mathematically, the difference in when we can effectively probe for each is notable. However, the fact that linear probes cannot decode parity in early latent representations does not mean this information is not present. It may simply be encoded via nonlinear structure.

One explanation might be that learning to extract the parity of a permutation is challenging because of how sensitive it is to the input representation. Parity is a canonical example of a sensitive function which transformers can have trouble learning \cite{hahn2024sensitive}. It is possible that allowing several rounds of generation makes this task easier for the model to learn. More speculatively, it may be that chain-of-thought offers additional benefits beyond increasing inference-time computation, enabling the model to learn functions that might be otherwise difficult to capture during training.

\subsection{Permutation encoding impacts when an answer becomes linearly decodable} 
The heatmaps in Figure \ref{fig:heatmap-order} are averaged over different permutation encodings. For a human mathematician, the choice of encoding can dramatically change how hard a task is. Calculating the length of a permutation is simple when one is handed a reduced Coxeter word; just count the number of generators. If one is handed a different encoding, additional computation may be required. Probing reveals similar patterns in PermuFormer. 

\begin{figure}[htbp]
    \centering
    \includegraphics[width=0.9\textwidth]{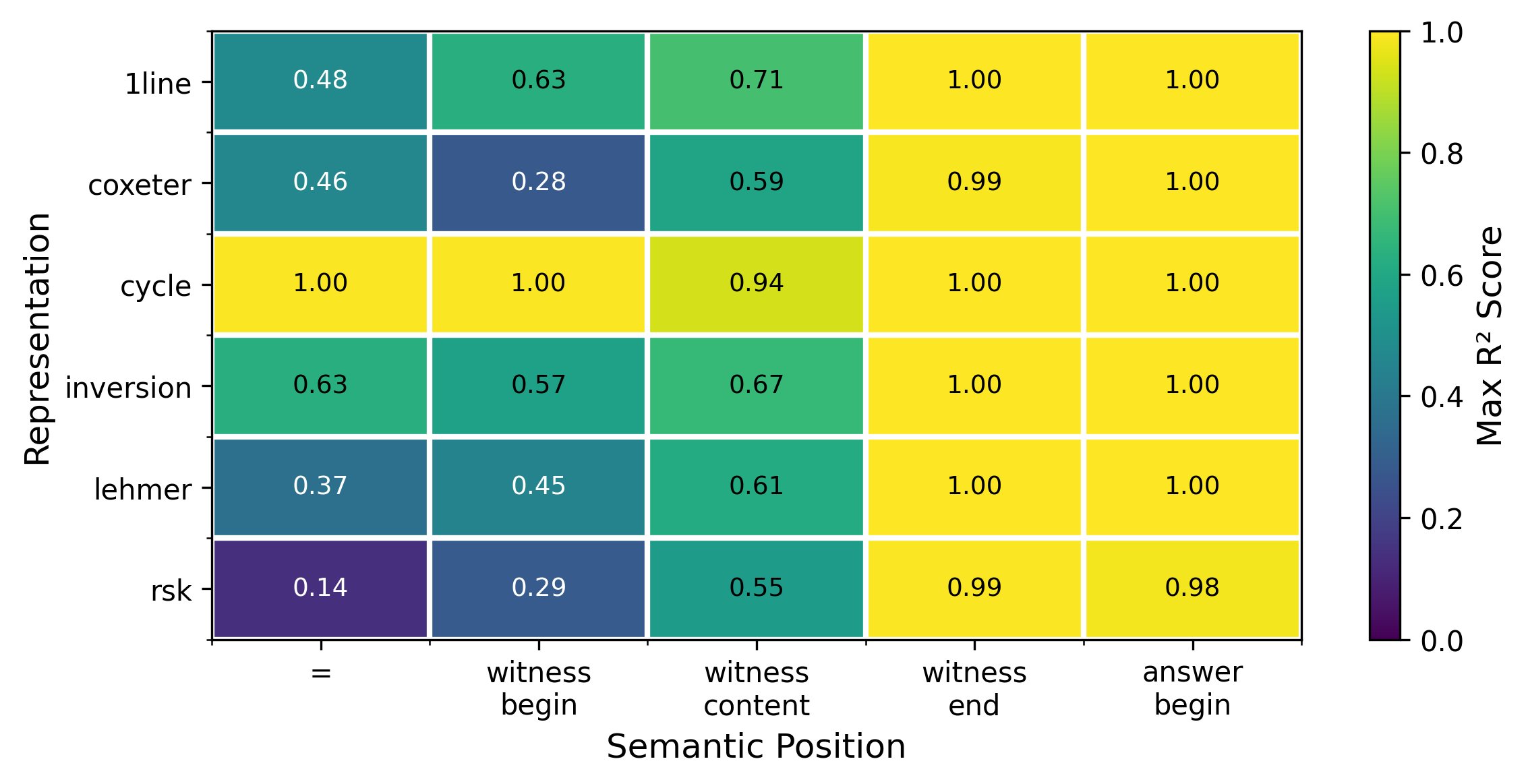}
    \includegraphics[width=0.9\textwidth]{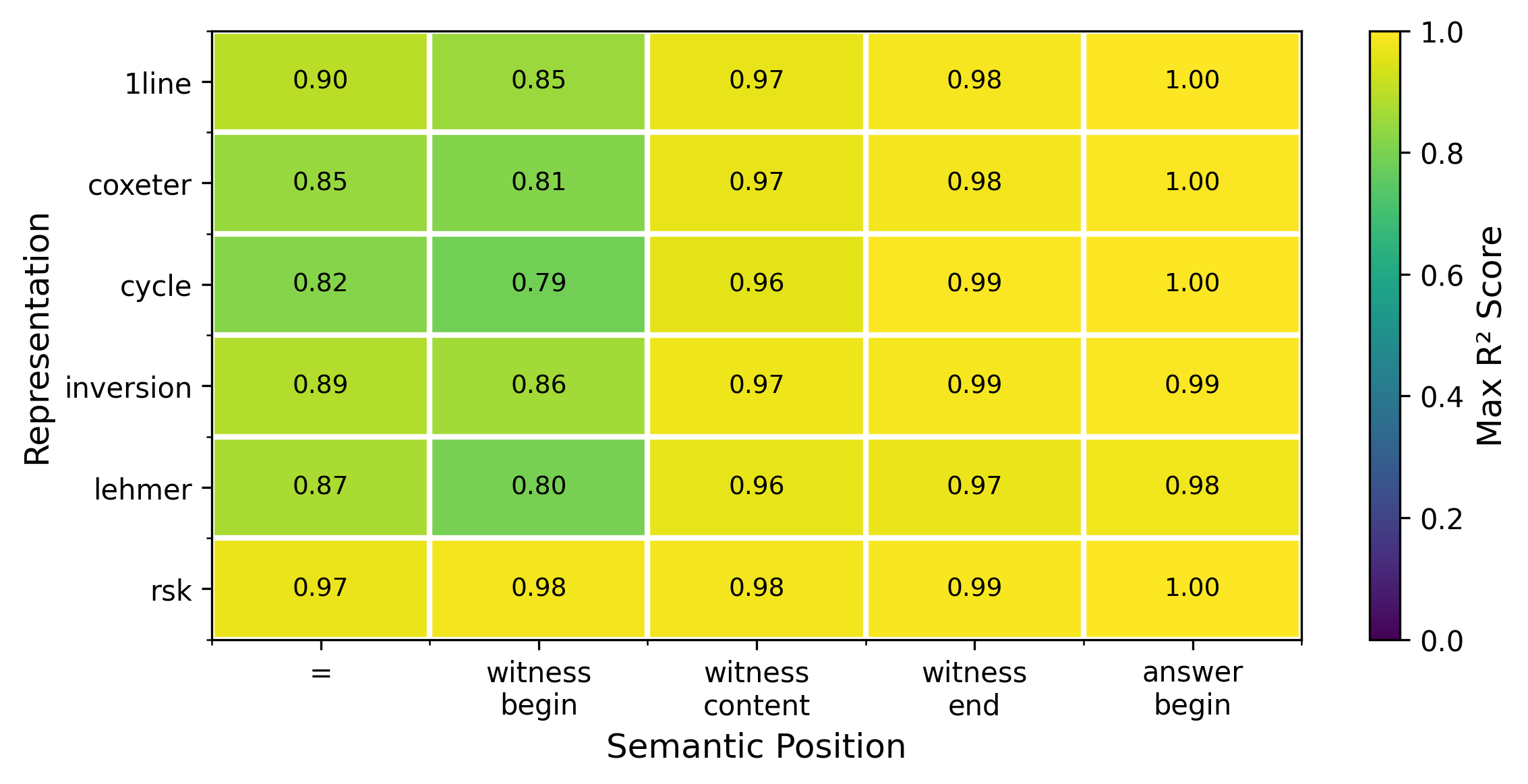}
    \caption{Heatmaps capturing probing $R^2$-scores across five phases of generation. \textbf{Top:} predicting the number of cycles in the permutation. \textbf{Bottom:} predicting the length of the longest increasing subsequence. (1) The final token of the prompt `='. (2) The delimiter marking the beginning of witness tokens. (3) The mean over witness token representations. (4) The delimiter marking the end of witness tokens. (5) The delimiter marking the beginning of the answer. The score shown is the maximum across all 12 blocks of the model.}
    \label{fig:heatmap-number-of-cycles}
\end{figure}

Figure \ref{fig:heatmap-number-of-cycles} (top) shows the maximum probe $R^2$ score (across blocks) for the task of predicting the number of cycles in a permutation. We break this down as a function of permutation encoding (rows) and stage of generation (columns). Note that calculating the number of cycles in a permutation is very easy if one is given the permutation in cycle notation. In this case one only needs to count the number of `(', `)' pairs. Additional computation is required for other encodings. We see this intuition reflected in the probing patterns in the figure. Probes achieve high $R^2$ scores on the initial representation of the prompt when the encoding is already in cycle notation, but this happens much later in generation when other encodings are used.

Another example can be found in the bottom heatmap in Figure \ref{fig:heatmap-number-of-cycles}. Here we show probes applied to the task of predicting the length of the longest increasing subsequence in a permutation. This statistic is easy to extract when the permutation is given in terms of its RSK representative (we only need to take the length of the first row in the common tableau shape). Again, encodings where this statistic is easily extractable for a human are also the encodings where probes achieve high predictive performance early in the generation process. These results suggest that, like humans, PermuFormer exploits the information available in an encoding to simplify the process of solving a task.

\subsection{Latent Space Visualization}
\label{sect:visualizations}

To gain better intuition for the way that PermuFormer represents permutations, we use UMAP \cite{mcinnes2018umap} to visualize activations. To stay in-distribution, we have to situate permutations within a specific task found in PermuFormer’s training distribution. We choose the task of inversion because here the input and output are encoding dependent. This helps to isolate encoding-specific aspects of the representation which might otherwise be obscured by convergence to a common output (e.g., if the task is predicting permutation length, then hidden activations will converge to the same length token regardless of encoding).

We choose to extract activations from the sixth block of the network, since prior work has found that the middle of a transformer is typically where semantic structure is observed \cite{skean2024does}. Finally, we look at the representation on the first generated token, which is the delimiter for the encoding type (e.g., \texttt{1linebegin} or \texttt{cyclenotationbegin}) in the inverse calculation task. We found that UMAP visualizations of activations of this token exhibited richer geometric structure than representations at both earlier and later stages of generation. A more thorough, mechanistic study would explore the dynamics across generated tokens, as this could lead to insights into how PermuFormer performs these calculations.

\begin{figure}[t]
    \centering

    \begin{subfigure}[t]{0.43\textwidth}
        \centering
        \includegraphics[width=\linewidth]{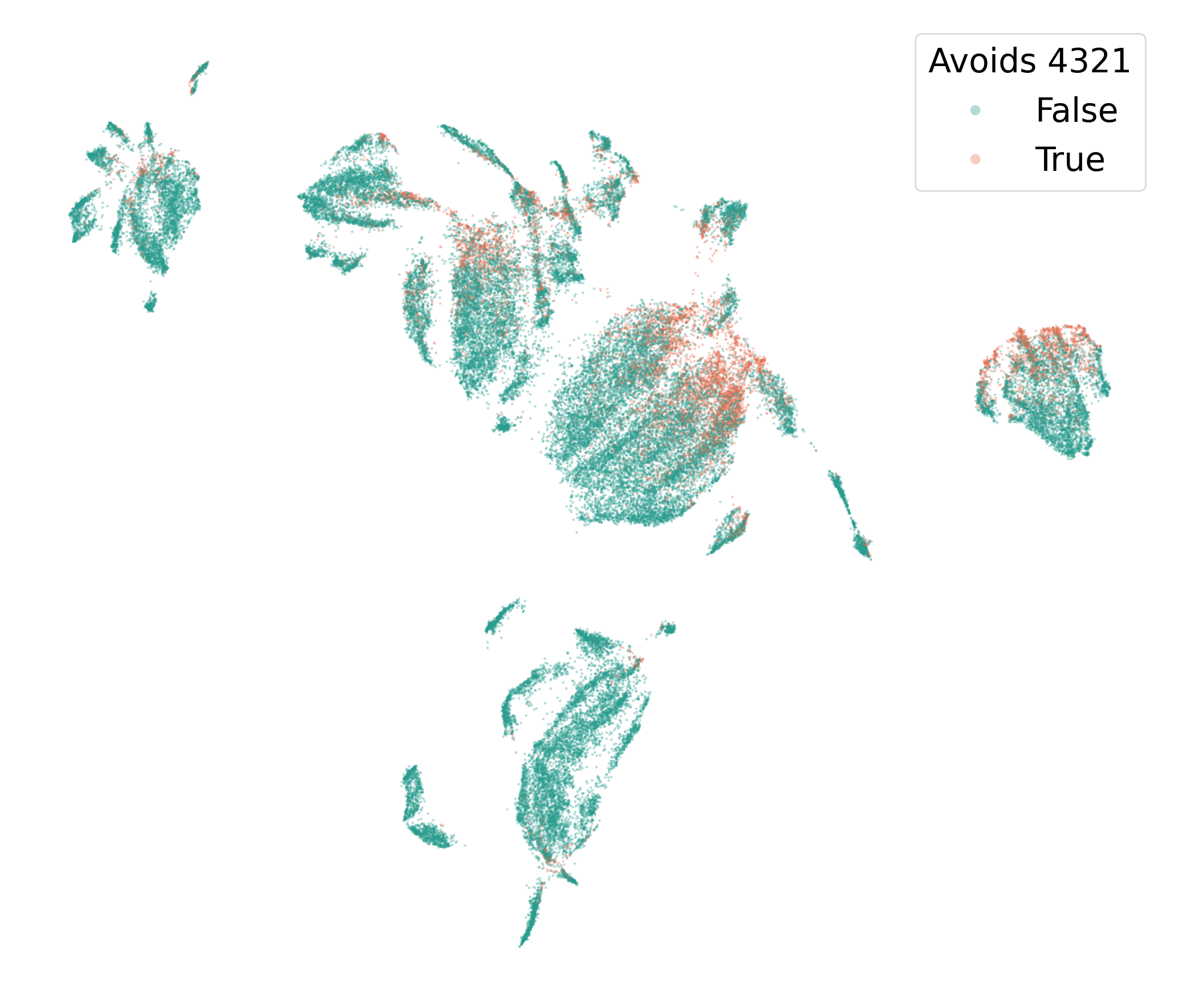}
    \end{subfigure}
    \hfill
    \begin{subfigure}[t]{0.43\textwidth}
        \centering
        \includegraphics[width=\linewidth]{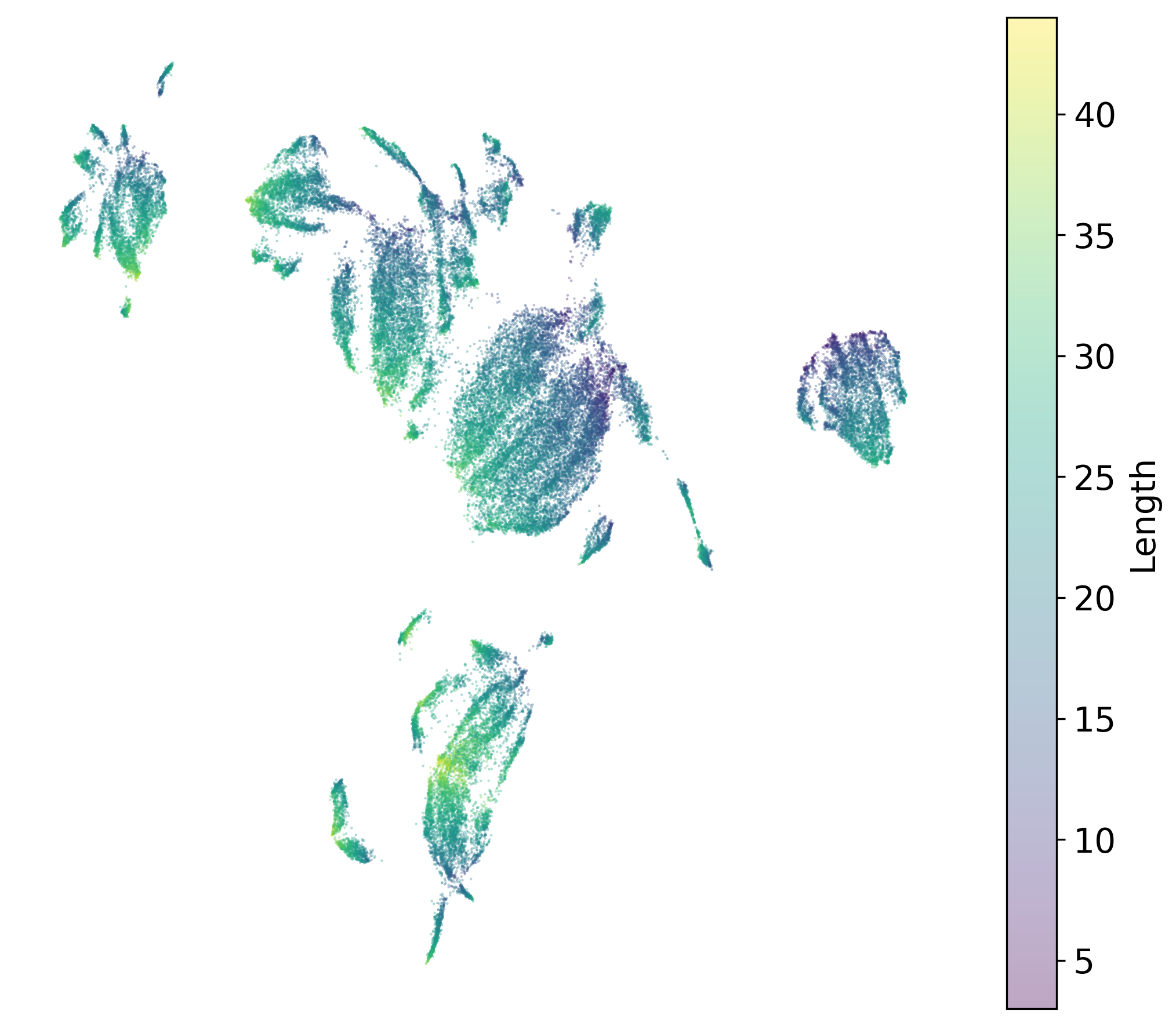}
    \end{subfigure}

    \vspace{0.8em}

    \begin{subfigure}[t]{0.43\textwidth}
        \centering
        \includegraphics[width=\linewidth]{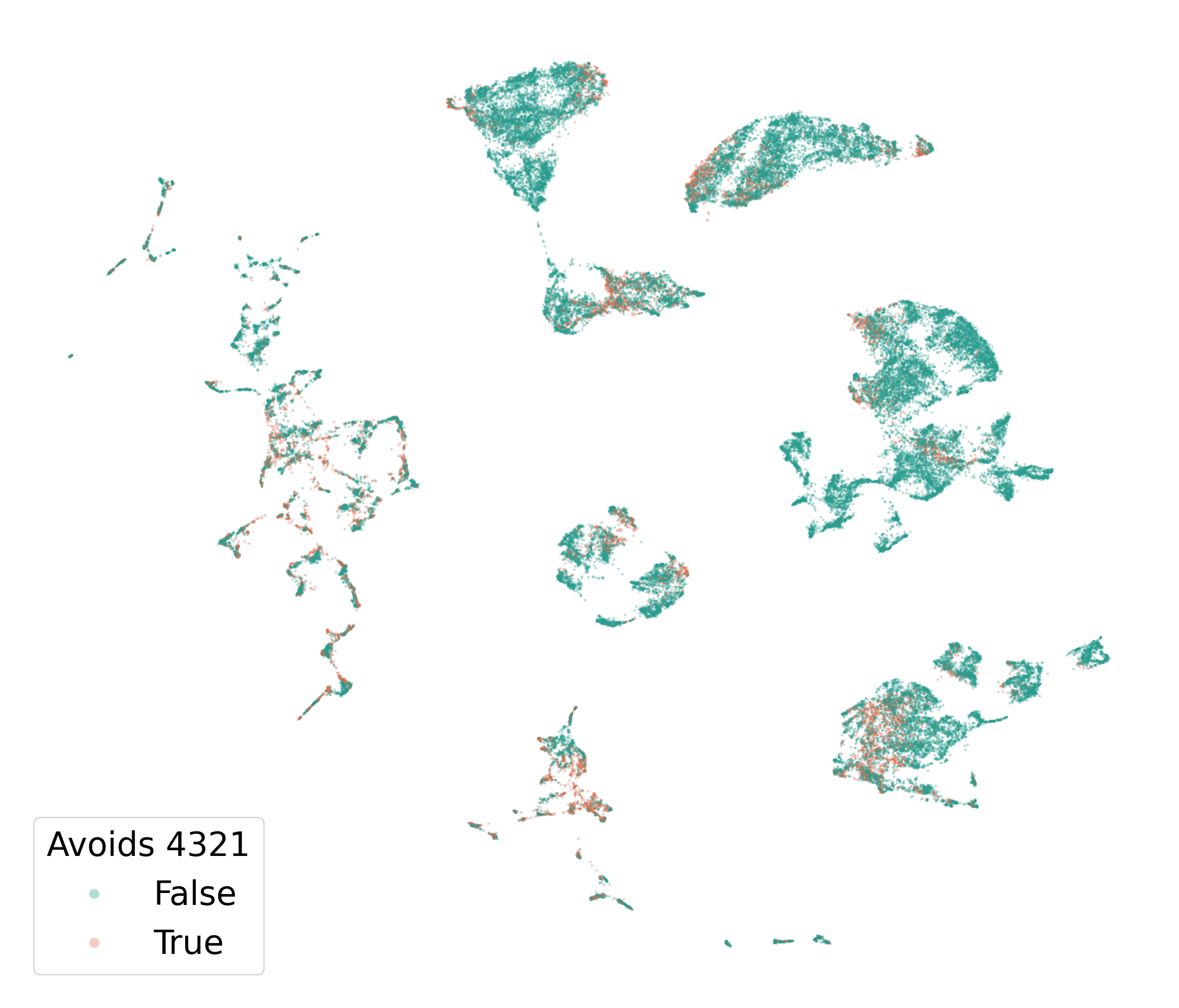}
    \end{subfigure}
    \hfill
    \begin{subfigure}[t]{0.43\textwidth}
        \centering
        \includegraphics[width=\linewidth]{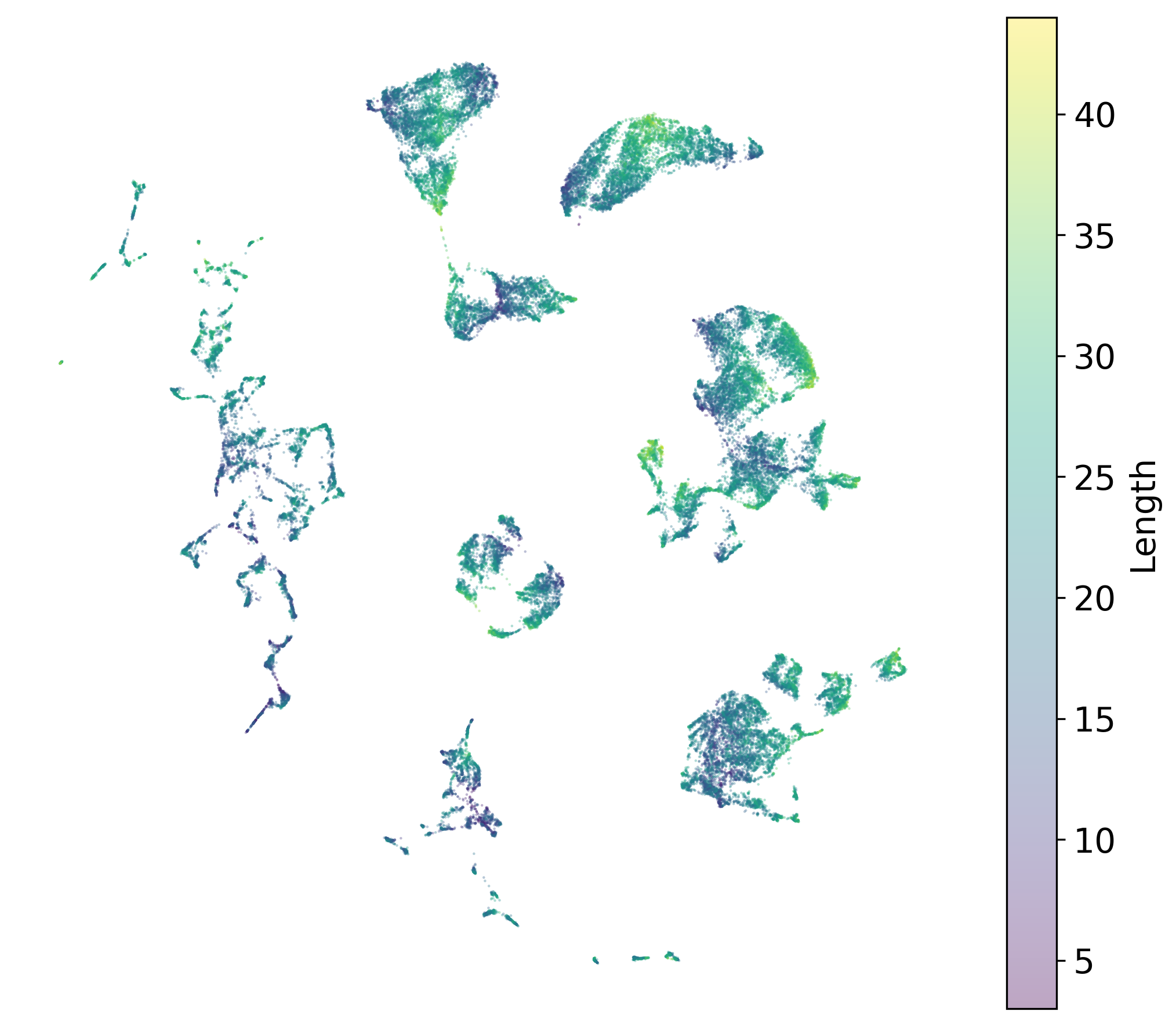}
    \end{subfigure}

    \vspace{0.8em}

    \begin{subfigure}[t]{0.43\textwidth}
        \centering
        \includegraphics[width=\linewidth]{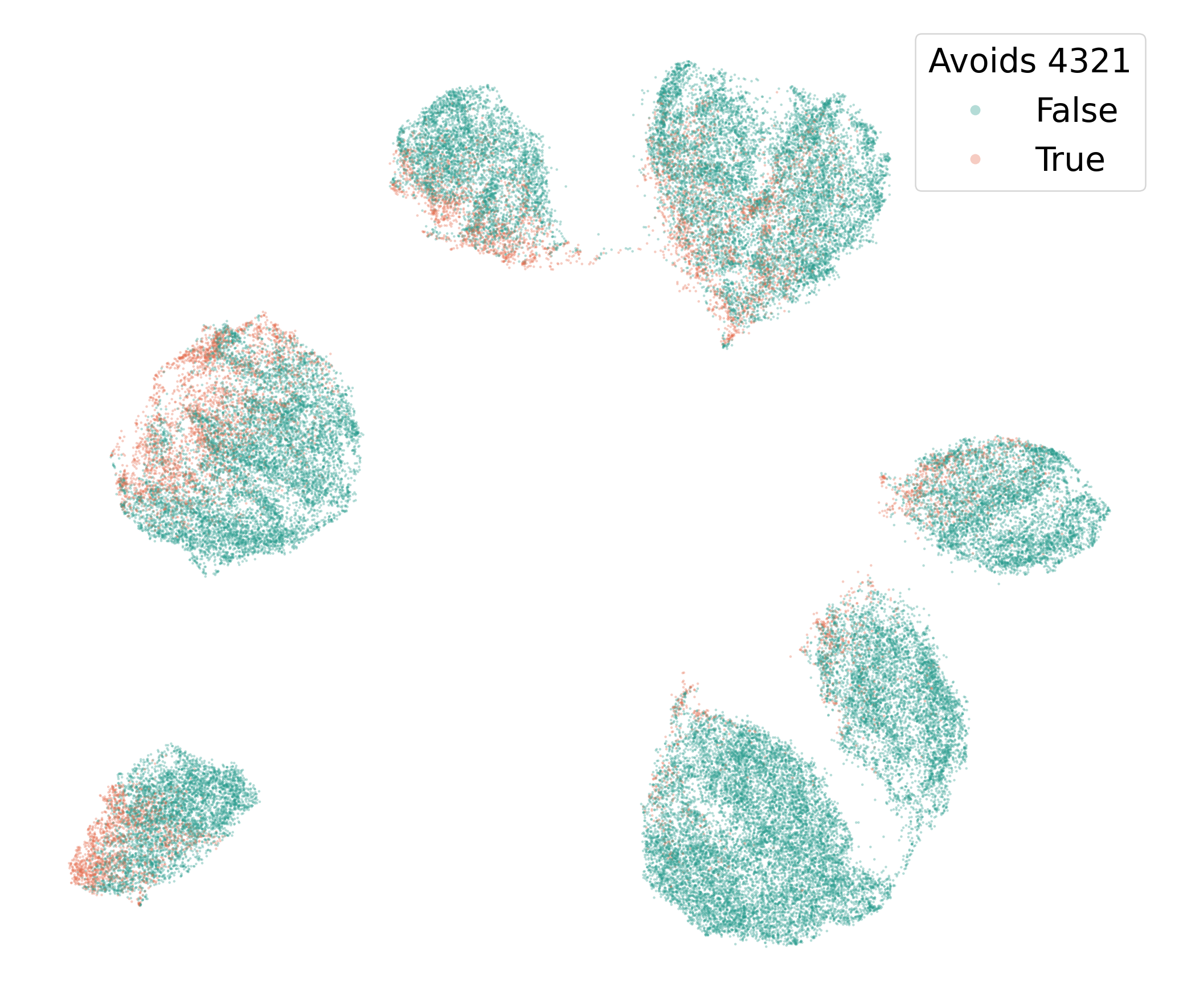}
    \end{subfigure}
    \hfill
    \begin{subfigure}[t]{0.43\textwidth}
        \centering
        \includegraphics[width=\linewidth]{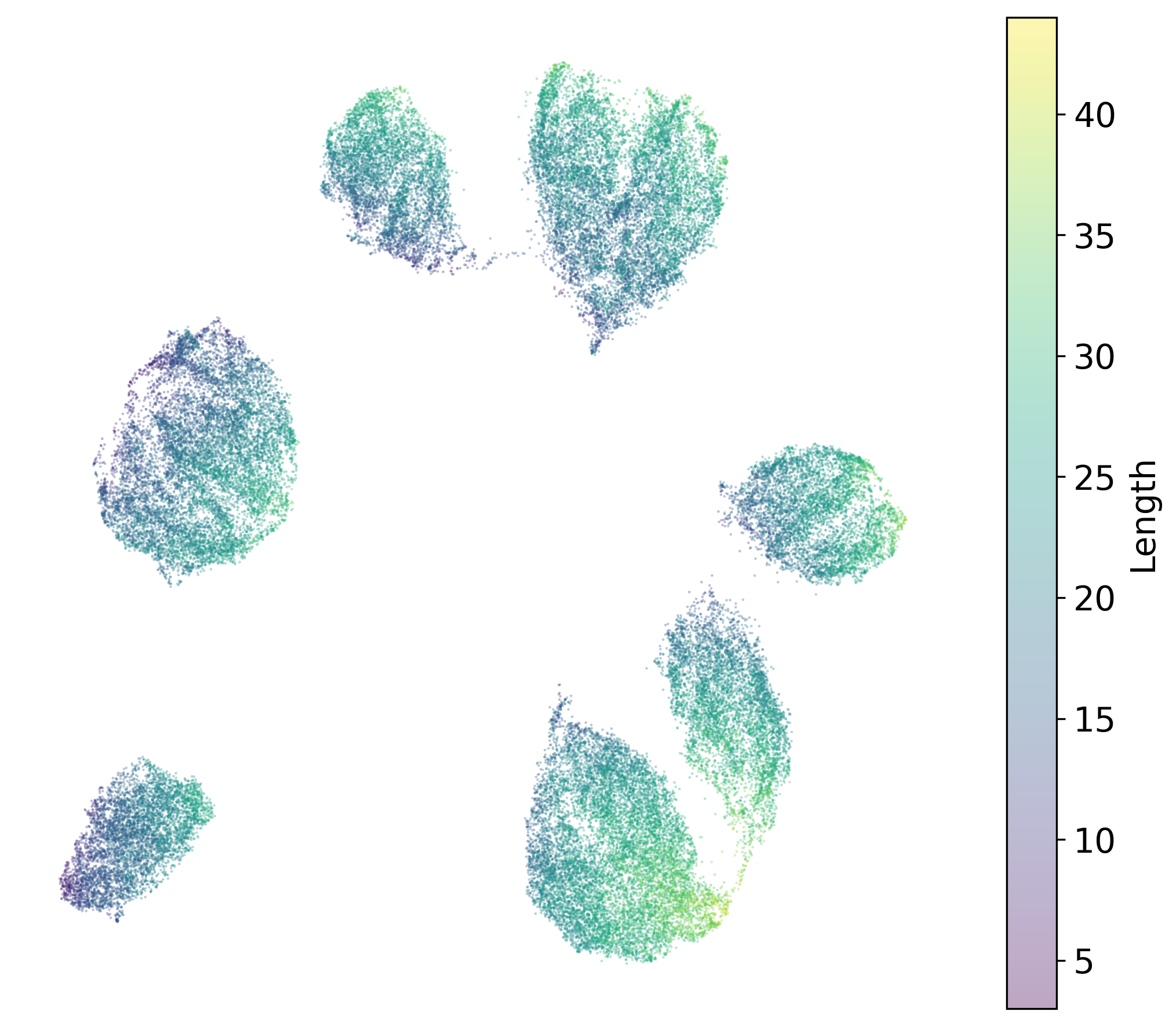}
    \end{subfigure}

    \caption{UMAP visualizations of activations corresponding to the first generated token for the task of permutation inversion. \textbf{Top row:} Permutations given in one-line notation. \textbf{Middle row:} Permutations given as a Coxeter reduced expression. \textbf{Bottom row:} Permutations given via Lehmer code. \textbf{First column:} Colored by whether or not a permutation avoids the pattern $4321$. \textbf{Second column:} Colored by the length of the permutation.}
    \label{fig:umap-visualizations}
\end{figure}

The UMAP visualizations can be found in Figure \ref{fig:umap-visualizations}. The first row corresponds to one-line notation, the second row corresponds to Coxeter reduced word, and the third row corresponds to Lehmer code. These were chosen because they illustrate some of the structural distinctions we see across encodings. The most striking feature is how different the visualizations appear, from the number of clusters to the apparent density. The visualization of permutations in one-line notation consists of a handful of distinct superclusters that can be further subdivided into smaller components. On the other hand, the visualization of permutations encoded as Coxeter reduced words is highly fragmented. Finally, the visualization of permutations written in Lehmer code consists of seven well-defined, uniform clusters. While it is easy to overinterpret UMAP, the visualizations are consistent with the idea that PermuFormer maintains encoding-dependent representations. 

To better understand the way that permutations are organized within these representations, we display the scatterplots with two different schemes for coloring the points. In the first column of Figure \ref{fig:umap-visualizations}, we color points by whether the permutation is $4321$-avoiding or not. In the second, we color points by Coxeter length. In both cases, permutations tend to be situated near other permutations with the same property value, but this is not what determines the global clustering in the visualization. We found this to be true across all the Boolean or integer-valued properties that we inspected. The only statistics we observed that were not spatially organized in the visualizations were permutation sign and permutation order, which is consistent with our probing results suggesting that these do not become linearly decodable until later in generation.

The origin of the large-scale structure observed in Figure \ref{fig:umap-visualizations} remains unclear but may be related to the specific task of calculating an inverse. Significant differences between encodings raise the possibility that PermuFormer exploits encoding-specific strategies for computing permutation inverses. We explore this in the next section.

\subsection{Does PermuFormer Learn a Unified Representation of Permutations?}

PermuFormer is trained to perform any task given any of the six different encodings of a permutation. The visualizations in Section \ref{sect:visualizations} suggest some similarities and differences between representations of different encodings. In this section, we continue this analysis with a specific focus on the question of whether PermuFormer learns a unified representation of permutations. 

Moving beyond visualization alone, we use centered kernel alignment (CKA) \cite{kornblith2019similarity} to compare the internal representations of permutations expressed in different encodings. Specifically, we fix a list of permutations from $S_9$. For each of these permutations $\sigma$, we generate a 6-tuple that consists of all six encodings of $\sigma$ (e.g., $\sigma$ written in one-line notation, $\sigma$ written in cycle notation, etc.). We can then insert these permutations into a chosen task to get a set of examples that are identical up to the permutation encoding.

\begin{figure}[htbp]
    \centering
    \includegraphics[width=0.9\textwidth]{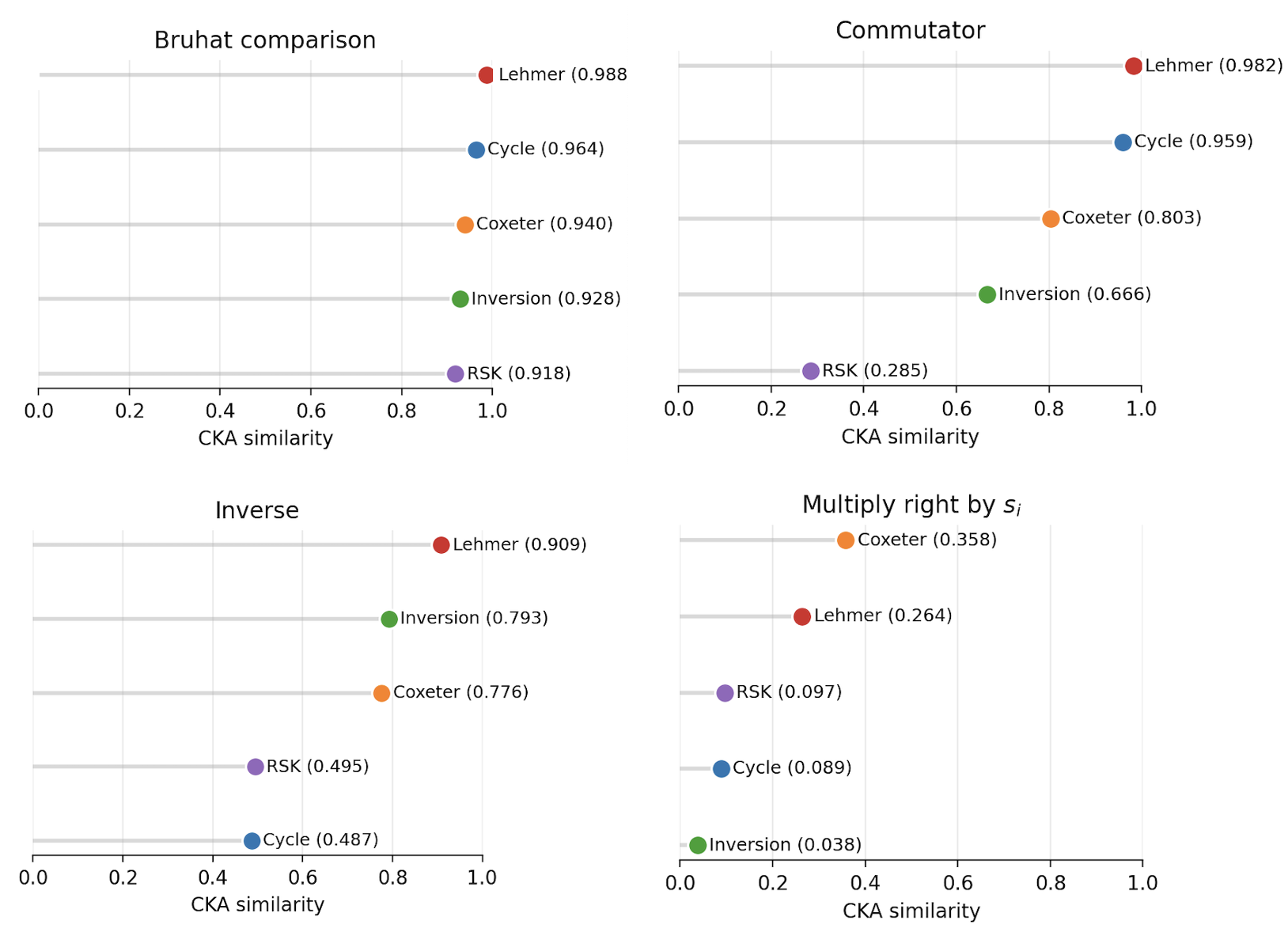} 
    \caption{CKA scores for different encodings relative to one-line notation. The $x$-axis corresponds to CKA score, and horizontal bars correspond to encodings. Each subfigure corresponds to a different task.}
    \label{fig:cka-plots}
\end{figure}

In Figure \ref{fig:cka-plots}, we show results of an experiment comparing representations of each encoding to the representation of one-line notation on four different tasks: Bruhat comparison, calculation of the commutator, calculation of the inverse, and multiplication by the transposition $s_i$ on the right. CKA scores are calculated over 100 examples at the sixth block and first generated token. We see that CKA scores differ substantially based on the task. When performing Bruhat comparison, CKA values between activations for one-line notation and the other encodings are close to one. On the other hand, when performing multiplication by an adjacent transposition $s_i$, CKA scores are less than 0.5\footnote{Note that for this task, Coxeter reduced word representations are more similar to one-line notation representations than any other encoding. This is interesting given that multiplication by adjacent transpositions is naturally expressed in this encoding.}. Beyond this, the relative order of encodings by their CKA similarity to one-line notation representations also differs from task to task. When calculating the inverse of a permutation, Lehmer code, inversion vector, and Coxeter reduced word encodings all have the highest CKA scores. When calculating the commutator, Lehmer code and cycle notation are the most similar.

This experiment is inconsistent with the simplistic hypothesis that PermuFormer learns a single universal representation for permutations that works for all encodings. On the other hand, the high CKA scores between encodings of other tasks support that idea that different encodings may lead to aligned representations. While further experiments are needed to provide more evidence, we conjecture that PermuFormer learns flexible task-specific representations that are aligned in some cases, while remaining distinct in others.

\section{Conclusion}

In this work, we described PermuFormer, a model with extensive multi-task, multi-encoding pretraining for permutation-focused tasks in algebraic combinatorics. PermuFormer is small and only requires a limited amount of compute for fine-tuning. We show that PermuFormer is an effective starting point for a range of tasks, from routine calculations in algebraic combinatorics like pattern avoidance to research-level tasks like the prediction of Schubert polynomial structure constants. We hope that this work will serve as a starting point for future research exploring methods of pretraining for transferable mathematical representations. While massive, frontier LLMs have been shown to have remarkable problem-solving skills, we believe scaling diverse, computation-based pretraining may lead to models that can offer us new types of mathematical insights, distinct from text-based models.

\begin{ack}
This work was conducted under the Laboratory Directed Research and Development Program at PNNL, a multi-program national laboratory operated by Battelle for the U.S. Department of Energy under contract DE-AC05-76RL01830.

The author would like to thank Helen Jenne, Sarah Scullen, Michael Zeng, and Junaid Hasan for useful discussions.
\end{ack}

\bibliographystyle{plain}
\bibliography{references}

\medskip


\newpage

\appendix

\section{Task details}
\label{sect:permutation-properties}

We begin this section by reviewing permutation properties not covered in Section \ref{sect-background}. For all examples, $\sigma$ is a permutation in $S_n$ and $i,j \in [n]$.

\begin{itemize}
    \item \textbf{Fixed points:} The \emph{fixed points} of a permutation $\sigma$ are elements $i$ such that $\sigma(i) = i$.
    \item \textbf{Descents:} A \emph{descent} in $\sigma$ is an element $i \in [n-1]$ where $\sigma(i) > \sigma(i+1)$.
    \item \textbf{Recoils:} An integer $i \in [n]$ with $i < n$ is a \emph{recoil} if $\sigma^{-1}(i+1) < \sigma^{-1}(i)$.
    \item \textbf{Peaks:} An integer $i \in [n]$ with $1 < i < n$ is a \emph{peak} if $\sigma(i-1) < \sigma(i)$ and $\sigma(i+1) < \sigma(i)$.
    \item \textbf{Longest increasing subsequences:} A \emph{length $k$ increasing subsequence} in $\sigma$ is a sequence of integers $i_1 < i_2 < \dots < i_k$ such that $\sigma(i_1) < \sigma(i_2) < \dots < \sigma(i_{k})$. Subsequences of $\sigma$ that achieve maximal length are called its \emph{longest increasing subsequences}.
    \item \textbf{Derangements:} A permutation is a \emph{derangement} if it has no fixed points (i.e., $\sigma(i) \neq i$ for all $i \in [n]$).
    \item \textbf{Number of cycles:} This is just the length of the cycle type. If $\sigma$ has cycle type $(\lambda_1,\lambda_2, \dots, \lambda_k)$, then the number of cycles is $k$.
    \item \textbf{Order:} The order of $\sigma$ is the smallest integer $k$ such that $\sigma^{k} = e$.
    \item \textbf{RSK shape:} The RSK algorithm maps a permutation to a pair of standard Young tableaux of the same shape, $(P,Q)$. The RSK shape of the permutation is the common shape of $P$ and $Q$.
    \item \textbf{Longest decreasing subsequence length:} This is analogous to the longest increasing subsequence. A decreasing subsequence of length $k$ is a sequence $i_1 < i_2 < \dots < i_k$ such that $\sigma(i_1) > \sigma(i_2) > \dots > \sigma(i_k)$. Such subsequences of $\sigma$ that achieve maximal length are called its \emph{longest decreasing subsequences}.
    \item \textbf{Involutions:} A permutation $\sigma$ is an involution if $\sigma^2 = e$.
    \item \textbf{Vexillary permutations:} These are permutations that avoid $2143$.
    \item \textbf{Grassmannian:} Grassmannian permutations are those that have at most one descent. That is, there is at most one $i \in [n-1]$ such that $\sigma(i) > \sigma(i+1)$.
    \item \textbf{Excedances:} These are indices $i$ such that $\sigma(i) > i$.
\end{itemize}

We also review permutation operations that we did not cover in Section \ref{sect-background}. Here, $\pi,\tau \in S_n$.

\begin{itemize}
    \item \textbf{Conjugation:} Conjugation of $\sigma$ by $\pi$ is $\pi \sigma \pi^{-1}$.
    \item \textbf{Relative left:} We use this terminology to denote $\sigma^(-1)\pi$.
    \item \textbf{Relative right:} We use this terminology to denote $\pi\sigma^(-1)$.
    \item \textbf{Commutator:} The commutator of $\sigma$ and $\tau$ is $\sigma^{-1}\tau^{-1}\sigma \tau$.
    \item \textbf{Power:} The $k$th power of $\sigma$ is $\sigma^k$ where $k$ is sampled uniformly from $\{2, 3, 4\}$.
    \item \textbf{Multiplication on the left by $s_i$:} This is $s_i \sigma$.
    \item \textbf{Multiplication on the right by $s_i$:} This is $\sigma s_i$.
    \item \textbf{Left descent test:} This takes $\sigma$ and $i$ and returns `true' if $i$ is a descent of $\sigma^{-1}$ and `false' if not.
    \item \textbf{Right descent test:} This takes $\sigma$ and $i$ and returns `true' if $i$ is a descent of $\sigma$ and `false' if not.
    \item \textbf{Complement:} This is the permutation that sends $i$ to $n + 1 - \sigma(i)$.
    \item \textbf{Reverse:} This is the permutation obtained by reversing the order of $\sigma$ written in one-line notation.
\end{itemize}

\section{Training data}

\subsection{Dataset generation}
\label{sect:dataset-generation}

As mentioned in Section \ref{sect:training_data}, training and evaluation data is generated for PermuFormer procedurally, using SageMath \cite{sagemath}. For each instance we start by sampling uniformly from the three task families. Having done this, we sample uniformly from all tasks within that family. In the case of translation tasks, this determines the encodings (since these are a part of the definition of a translation task). For property extraction and operation tasks, we uniformly sample an encoding type. Finally, we select the permutation that will be used in the instance by sampling an integer $n$ from $\{2, \dots, 11\}$ and then generating a random ordering of $1, 2, \dots, n$. In cases where the task involves two permutations (like composition), we use the same $n$ but perform the random re-ordering twice. 

Note that because there is equal probability of generating an example from each family but these families have different numbers of corresponding tasks, the probability of drawing different individual tasks can differ. This is reflected in the task counts found in Tables \ref{table:property-volume}, \ref{table:operation-volume}, and \ref{table:translation-volume}. A similar phenomenon appears in terms of the number of permutations of a given size. If we were to simply sample uniformly across the union of all symmetric groups $S_2, \dots, S_{11}$ we would almost always get a permutation of size 10 or 11. Ensuring that we sample uniformly across all permutation sizes introduces oversampling of smaller permutations. However, oversampling was necessary to ensure PermuFormer had exposure to tasks using smaller permutations.

The dataset is split into 99.5\% training data and 0.5\% evaluation data. We sampled evaluation data first and then checked every prospective training instance to ensure that it was not in the evaluation set. An example is considered a duplicate if the task, encoding, and permutations all match. Deduplication of the training set occurs via a rolling 5 million element window. 

\subsection{Tasks with multiple correct answers}
\label{sect:multiple-correct-answers}

Some of the tasks have multiple mathematically correct expressions within the syntax we have defined. For example, there are generally multiple reduced expressions for a permutation and a list of permutation fixed points can be listed in any order. We generally defaulted to the conventions used by Sage. These are:
\begin{itemize}
    \item The fixed points, peaks, recoils, and descents are all listed in increasing order.
    \item Cycle type is listed in decreasing order (as is the mathematical convention).
    \item The longest increasing subsequences are listed in reverse lexicographic order (e.g., \verb|[1,2,6]| comes before \verb|[1,2,5]|).
    \item The order of cycles follows the convention that the cycle containing $1$ is listed first. Then we list the cycle with the smallest element not already listed, etc. Cycles start with the smallest element in the cycle.
    \item We use the \verb|reduced_word()| command in Sage to generate reduced words. This generates a reduced word by converting the permutation into its Lehmer code and then working through this to generate a product of Coxeter generators with minimal length. Because this process is deterministic, the model sees only one canonical reduced word for a permutation. This is important to keep in mind when interpreting the results.
\end{itemize}

\begin{table}
  \caption{Permutation property tasks along with their count in the training and evaluation sets.}
  \label{table:property-volume}
  \centering
  \begin{tabular}{lll}
    \toprule
    Property     & \# in train     & \# in test  \\
    \midrule
    Is derangement? & 420,908 & 2,148 \\
    Is even? & 417,635 & 2,166 \\
    Is vexillary? & 417,558 & 2,206 \\
    Is involution? & 419,435 & 2,098 \\
    Is Grassmannian? & 416,748 & 2,182 \\
    Cycle type & 420,635 & 2,159 \\
    Order & 420,382 & 2,241 \\
    Longest increasing subsequences (LIS) & 419,994 & 2,140 \\
    LIS length & 419,384 & 2,186 \\
    Longest decreasing subsequence (LDS) length & 420,028 & 2,154 \\
    Fixed points & 419,632 & 2,144 \\
    RSK shape & 419,770 & 2,165 \\
    Major index & 418,367 & 2,111 \\
    Descents & 417,138 & 2,110 \\
    Peaks & 416,887 & 2,134 \\
    Sign & 416,251 & 2,086 \\
    Recoils & 415,867 & 2,147 \\
    Length & 415,696 & 2,125 \\
    \# of cycles & 420,752 & 2,179 \\
    \# Fixed points & 419,556 & 2,119 \\
    \# Inversions & 419,352 & 2,111 \\
    \# Descents & 416,678 & 2,185 \\
    \# of excedances & 416,367 & 2,144 \\
    Avoids 1324 & 418,781 & 2,147 \\
    Avoids 1234 & 418,665 & 2,220 \\
    Avoids 2413 & 416,848 & 2,175 \\
    Avoids 4321 & 416,258 & 2,083 \\
    Avoids 3412 & 415,534 & 2,179 \\
    Avoids 213 & 418,271 & 2,142 \\
    Avoids 312 & 417,611 & 2,117 \\
    Avoids 132 & 416,994 & 2,164 \\
    Avoids 321 & 416,885 & 2,134 \\
    \bottomrule
  \end{tabular}
\end{table}

\begin{table}
    \caption{Permutation operation and comparison tasks along with their count in the training and evaluation sets.}
    \label{table:operation-volume}
    \centering
    \begin{tabular}{lll}
        \toprule
        Operation & \# in train & \# in test \\
        \midrule
        Conjugation & 1,280,038 & 5,401 \\
        Product & 1,280,404 & 5,354 \\
        Relative left & 1,279,740 & 5,341 \\
        Relative right & 1,281,580 & 5,278 \\
        Bruhat comparison & 1,282,206 & 5,268 \\
        Commutator & 1,280,829 & 5,168 \\
        Power & 980,641 & 4,934 \\
        Left descent test & 1,028,002 & 4,931 \\
        Right descent test & 1,028,404 & 4,877 \\
        Multiply on the left by $s_i$ & 1,027,022 & 4,895 \\
        Multiply on the right by $s_i$ & 1,025,894 & 4,890 \\
        Complement & 881,121 & 4,742 \\
        Inverse & 875,203 & 4,681 \\
        Reverse & 881,221 & 4,667 \\
        \bottomrule
    \end{tabular}
\end{table}

\begin{table}
    \caption{Permutation translation tasks along with their count in the training and evaluation sets.}
    \label{table:translation-volume}
    \centering
    \begin{tabular}{llll}
        \toprule
        Source encoding & Target encoding & \# in train & \# in test \\
        \midrule
        one-line & Reduced Coxeter & 366,754 & 2,054 \\
        one-line & Cycle & 366,561 & 2,060 \\
        one-line & Inversion vector & 367,392 & 2,057 \\
        one-line & Lehmer & 366,586 & 1,999 \\
        one-line & RSK & 366,802 & 2,038 \\
        ReducedCoxeter & one-line & 367,771 & 1,902 \\
        Reduced Coxeter & Cycle & 367,565 & 2,042 \\
        Reduced Coxeter & Inversion vector & 366,618 & 2,069 \\
        ReducedCoxeter & Lehmer & 366,526 & 2,024 \\
        Reduced Coxeter & RSK & 366,589 & 2,020 \\
        Cycle & one-line & 366,432 & 2,025 \\
        Cycle & Reduced Coxeter & 367,474 & 1,931 \\
        Cycle & Inversion vector & 367,941 & 2,051 \\
        Cycle & Lehmer & 366,893 & 2,092 \\
        Cycle & RSK & 366,270 & 2,088 \\
        Inversion vector & one-line & 367,131 & 1,999 \\
        Inversion vector & Reduced Coxeter & 367,063 & 1,984 \\
        Inversion vector & Cycle & 367,135 & 2,062 \\
        Inversion vector & Lehmer & 368,073 & 2,051 \\
        Inversion vector & RSK & 366,027 & 2,101 \\
        Lehmer & one-line & 367,202 & 1,958 \\
        Lehmer & Reduced Coxeter & 366,367 & 2,032 \\
        Lehmer & Cycle & 366,502 & 2,022 \\
        Lehmer & Inversion vector & 366,069 & 1,933 \\
        Lehmer & RSK & 366,357 & 2,092 \\
        RSK & one-line & 367,647 & 2,012 \\
        RSK & Reduced Coxeter & 366,826 & 1,988 \\
        RSK & Cycle & 366,195 & 2,017 \\
        RSK & Inversion vector & 367,412 & 1,990 \\
        RSK & Lehmer & 366,648 & 2,080 \\
        \bottomrule
    \end{tabular}
\end{table}

\begin{table}
    \caption{A list of all tasks that use computational witnesses along with the nature of the witness.}
    \label{table:witnesses}
    \centering
    \begin{tabular}{ll}
        \toprule
        Task & Witness  \\
        \midrule
        Length & Reduced word \\
        \# of descents & List of descents \\
        \# of fixed points & List of fixed points \\
        \# of cycles & Cycle type \\
        \# of inversions & Inversion vector\\
        \# of excedances & List of excedances \\
        Sign & Length \\
        LIS length & RSK shape \\
        LDS length & RSK shape \\
        Major index & List of descents \\
        Is even? & Length \\
        Is derangement? & List of fixed points \\
        Is involution? & Inverse \\
        Is Grassmannian? & List of descents \\
        Is vexillary? & Examples of 2143 patterns \\
        Avoids $i_1\dotsi_k$ & Examples of $i_1\dots i_k$ patterns\\
        Order & Cycle type\\
        \bottomrule
    \end{tabular}
\end{table}

\begin{figure}[htbp]
    \centering
    \includegraphics[width=0.75\textwidth]{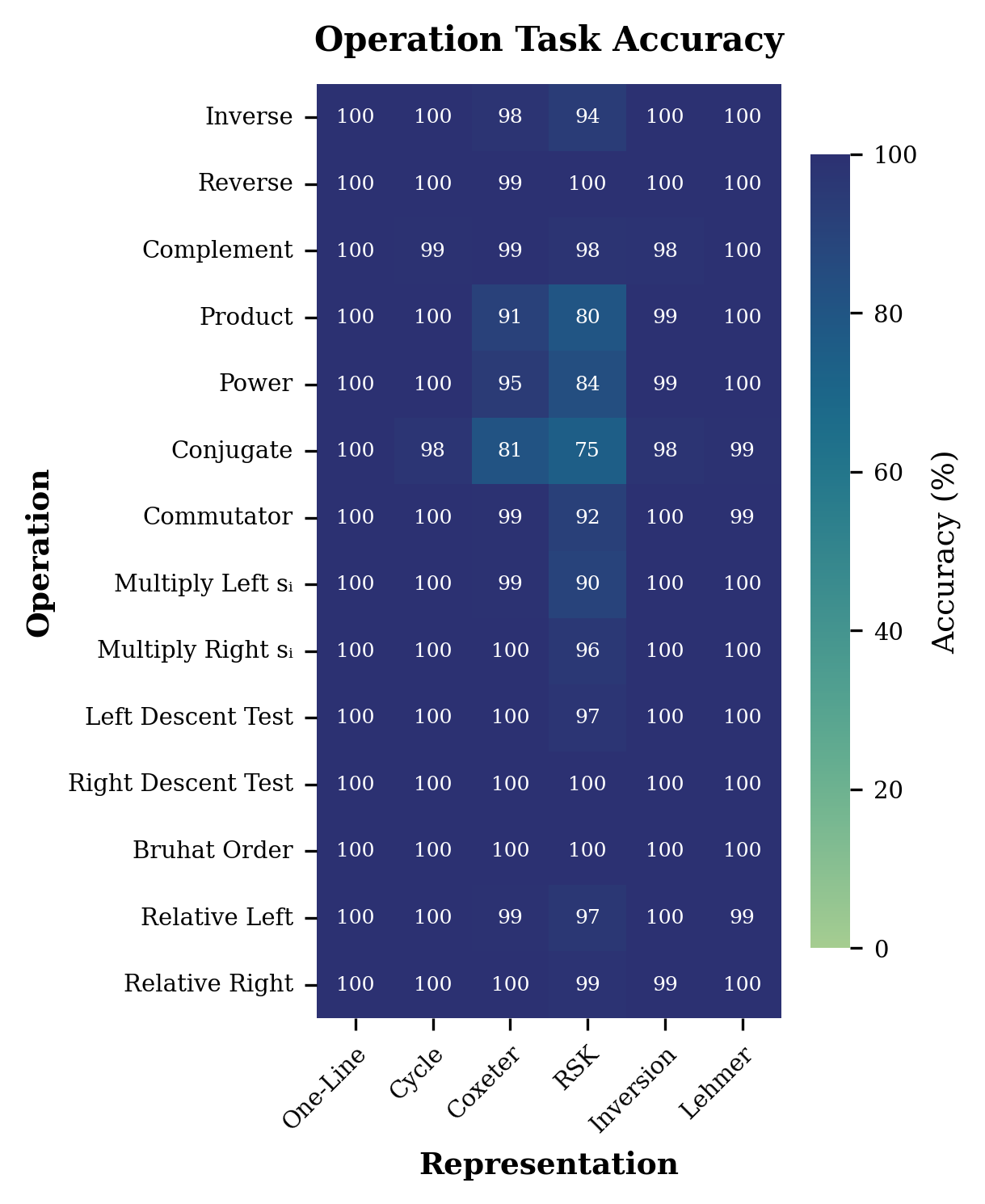}
    \caption{The evaluation accuracy across operation tasks as a function of the encoding of the permutation.}
    \label{fig:accuracy_across_operations}
\end{figure}

\begin{figure}[htbp]
    \centering
    \includegraphics[width=0.7\textwidth]{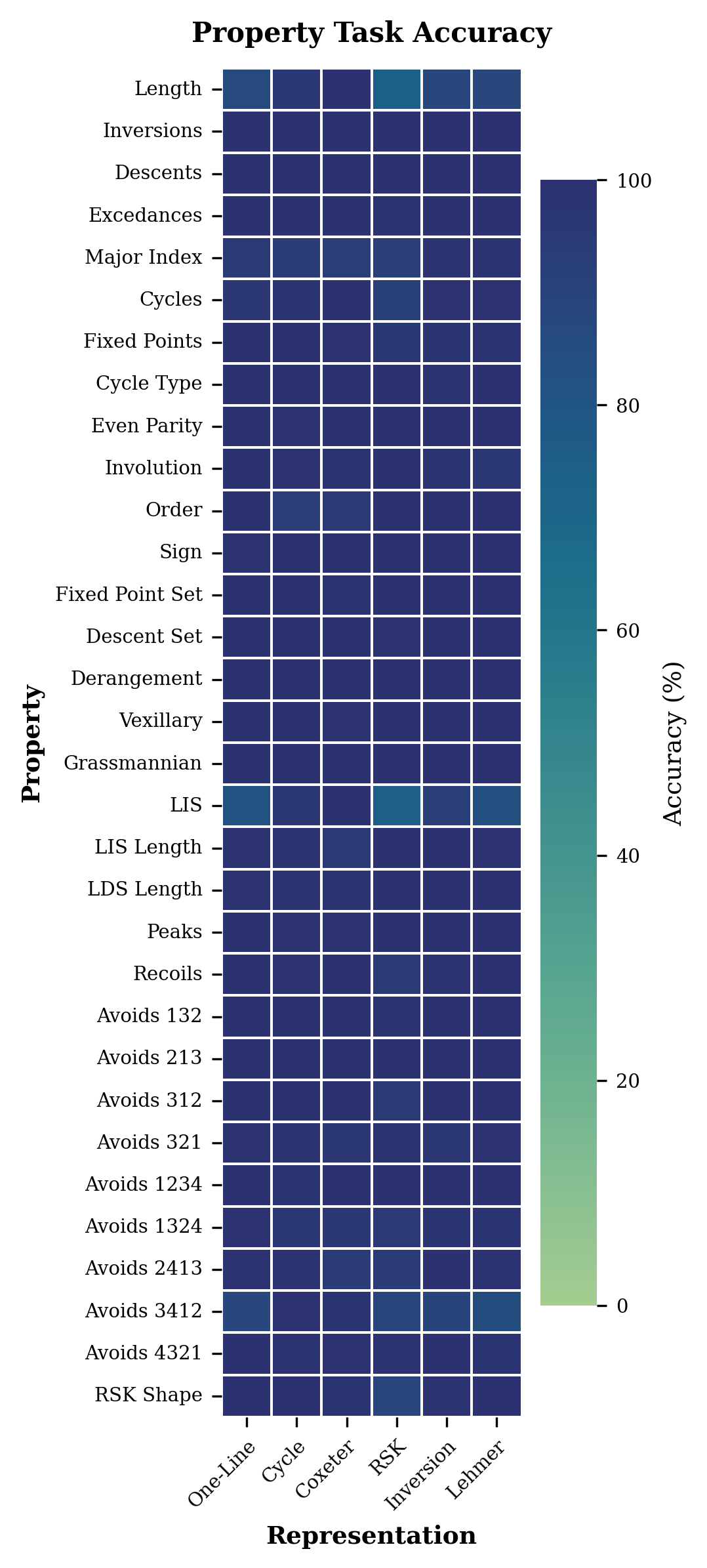}
    \caption{The evaluation accuracy across property extraction tasks as a function of the encoding of the permutation.}
    \label{fig:accuracy_across_properties}
\end{figure}

\begin{figure}[htbp]
    \centering
    \includegraphics[width=0.7\textwidth]{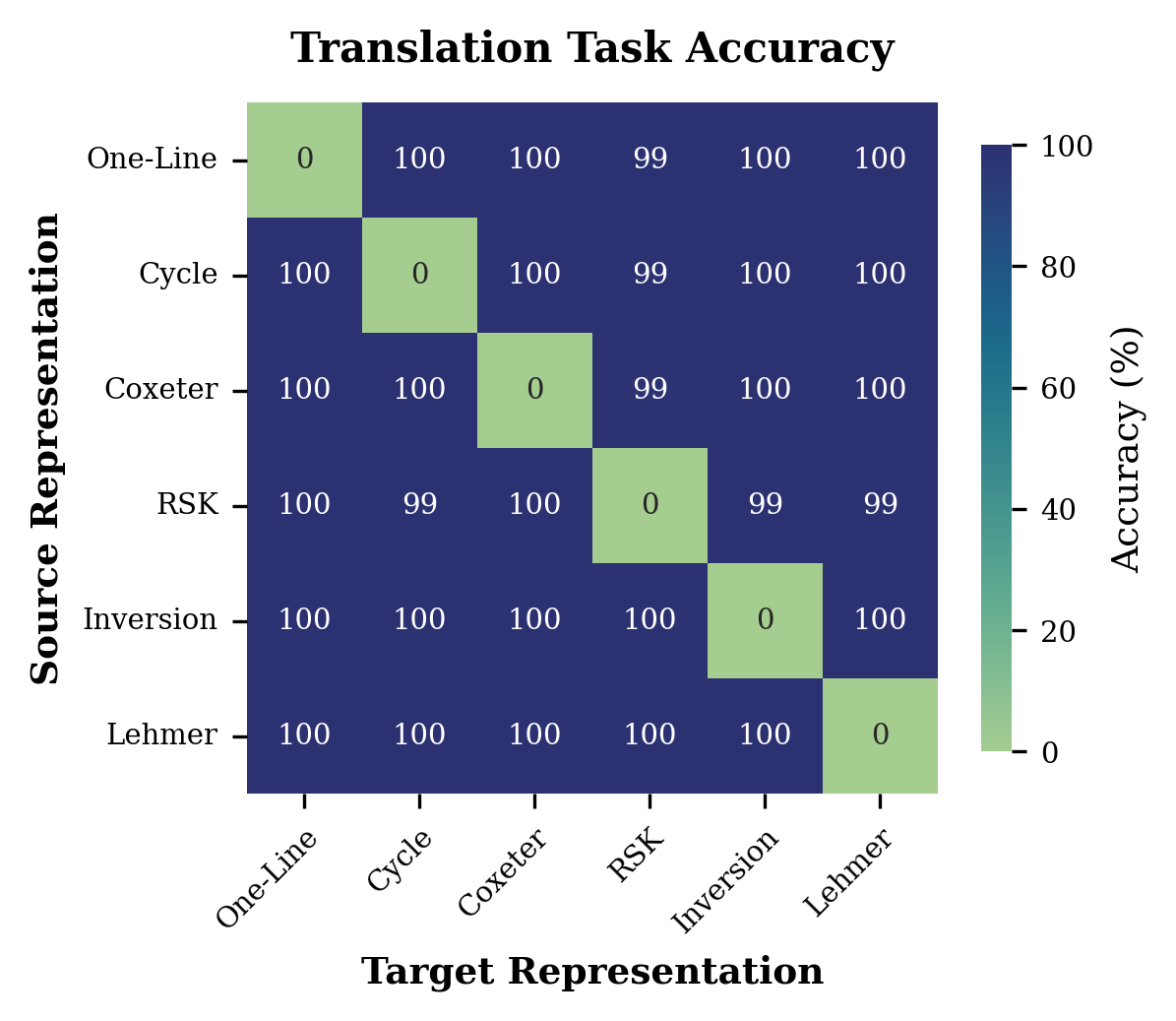}
    \caption{The evaluation accuracy for each translation task.}
    \label{fig:accuracy_across_translation}
\end{figure}

\section{Prompt for Frontier Model Comparison}
\label{sect:llm_prompt}

\begin{verbatim}
    
# Permutation Vocabulary Guide

This document describes a vocabulary used to represent permutations 
and operations.

## Core Principle

All data is tokenized as **space-separated strings** where numbers are kept 
as single tokens (e.g., `"11"` not `"1" "1"`). Each example is wrapped with 
`<|endoftext|>` tokens and begins with a size indicator token 
`n{size}` that specifies which symmetric group is 
being used (e.g., `n2` for S_2, `n7` for S_7, `n11` for S_{11}).

## Representation Formats

Permutations are encoded in six different representations, each with 
begin/end delimiters:

### 1. One-Line Notation
- **Format**: `1linebegin [ 2 , 3 , 1 ] 1lineend`
- **Meaning**: Standard array notation [\sigma(1), \sigma(2), \sigma(3), 
...]
- **Example**: `[2,3,1]` means 1→2, 2→3, 3→1

### 2. Cycle Notation
- **Format**: `cyclenotationbegin ( 1 , 2 , 3 ) cyclenotationend`
- **Meaning**: Disjoint cycle decomposition
- **Example**: `(1,2,3)` means 1→2→3→1

### 3. Coxeter Reduced Expression
- **Format**: `Coxeterreducedexpressionbegin [ 2 , 1 ] 
Coxeterreducedexpressionend`
- **Meaning**: Sequence of simple transpositions s_i (swap i and i+1)
- **Example**: `[2,1]` means s_2s_1 (first swap 1-2, then swap 2-3)

### 4. RSK Tableaux
- **Format**: `RSKtableauxbegin [ [ [ 1 ] , [ 2 ] ] , [ [ 1 ] , [ 2 ] ] ] 
RSKtableauxend`
- **Meaning**: Robinson-Schensted-Knuth correspondence [P-tableau, 
Q-tableau]
- **Structure**: Two standard Young tableaux as nested lists

### 5. Inversion Vector
- **Format**: `Inversionvectorbegin [ 2 , 1 , 0 ] Inversionvectorend`
- **Meaning**: Entry i counts inversions (j > i with \sigma(j) < \sigma(i))
- **Example**: For [3,2,1]: vector is [2,1,0]

### 6. Lehmer Code
- **Format**: `Lehmercodebegin [ 2 , 1 , 0 ] Lehmercodeend`
- **Meaning**: Factoradic representation (counts smaller elements 
to the right)

## Task Types

### Translation Tasks
Convert between representations using `in [target]make`:

```
1linebegin [ 2 , 1 ] 1lineend in cyclenotationmake =
cyclenotationbegin ( 1 , 2 ) cyclenotationend
```

### Property Tasks
Compute properties using `property [property]make`. Format includes 
**witnesses** (intermediate steps) and the final answer:

```
1linebegin [ 2 , 1 , 3 ] 1lineend property lengthmake =
witnessbegin [ 2 , 1 ] witnessend lengthbegin 1 lengthend
```

**Property Operations**:
- `lengthmake` → Coxeter length (minimum transpositions)
- `descentsmake` → Descent positions [i where \sigma(i) > \sigma(i+1)]
- `fixedpointsmake` → Fixed points [i where \sigma(i) = i]
- `isevenmake` → True/False (even permutation)
- `signmake` → +1 (even) or -1 (odd)
- `avoidsmake[pattern]` → Pattern avoidance (e.g., `avoidsmake[3,2,1]`)
- `cycletypemake` → Partition of cycle lengths
- `rskshapemake` → Young diagram shape
- `majorindexmake` → Sum of descent positions
- `numinversionsmake` → Number of inversions
- `ordermake` → Smallest k where \sigma^k = identity

**Witness Format**:
- `witnessbegin [data] witnessend` → Shows computation steps
- `witnessbegin nopattern witnessend` → Pattern not found
- For pattern avoidance: witness shows positions where pattern occurs 
(or `nopattern`)

### Algebraic Operations
Perform group operations:

**Binary Operations**:
```
1linebegin [ 2 , 1 ] 1lineend timesmake 1linebegin [ 2 , 1 ] 1lineend =
1linebegin [ 1 , 2 ] 1lineend
```

**Unary Operations**:
```
1linebegin [ 2 , 1 ] 1lineend inversemake =
1linebegin [ 2 , 1 ] 1lineend
```

**Operation Tokens**:
- `timesmake` → Product (composition)
- `inversemake` → Inverse permutation
- `powermake k` → k-th power (k \in {2,3,4})
- `conjugatemake` → Conjugation xyx^{-1}
- `commutatormake` → Commutator x^{-1}y^{-1}xy
- `relativeleftmake` → x^{-1}y
- `relativerightmake` → yx^{-1}
- `leftsiamake i` → Left multiply by s_i
- `rightsiamake i` → Right multiply by s_i
- `leftdescenttestmake i` → Is i a left descent?
- `rightdescenttestmake i` → Is i a right descent?
- `leftBruhatmake` → Bruhat order relation (x \leq y)
- `complementmake` → Complement: \sigma(i) → n+1-\sigma(i)
- `reversemake` → Reverse one-line notation

**Witnesses for Operations**:
- Bruhat order: `witnessbegin [[chain of permutations]] witnessend` or 
`witnessbegin nochain witnessend`

## Special Tokens

- `<|endoftext|>` → Example boundaries
- `n{size}` → 
Permutation size indicator (e.g., `n2`, `n7`, `n11`), appears immediately 
after opening `<|endoftext|>`
- `=` → Separates input from output
- `witnessbegin ... witnessend` → Computational trace
- `nopattern` → Single token indicating pattern not found 
(pattern avoidance)
- `nochain` → Single token indicating no Bruhat chain exists (Bruhat order)

## Tokenization Rules

1. **Numbers**: Single tokens (`11` not `1 1`)
2. **Booleans**: Single tokens (`True`, `False`)
3. **Structure**: Brackets, commas, parentheses are separate tokens
4. **Lists**: `[ item1 , item2 , item3 ]`
5. **Tuples**: `( item1 , item2 , item3 )`

## Example Complete Tasks

**Translation**:
```
<|endoftext|> n3 1linebegin [ 3 , 1 , 2 ] 1lineend in 
cyclenotationmake = cyclenotationbegin ( 1 , 3 , 2 ) 
cyclenotationend <|endoftext|>
```

**Property (with witness)**:
```
<|endoftext|> n3 1linebegin [ 3 , 2 , 1 ] 1lineend property 
avoidsmake [ 3 , 2 , 1 ] = witnessbegin [ 1 , 2 , 3 ] 
witnessend avoidsbegin False avoidsend <|endoftext|>
```

**Operation**:
```
<|endoftext|> n3 1linebegin [ 2 , 1 , 3 ] 1lineend inversemake 
= 1linebegin [ 2 , 1 , 3 ] 1lineend <|endoftext|>
```

## Key Insights for LLMs

1. Every example starts with `<|endoftext|>` followed immediately 
by `n{size}` indicating the symmetric group
2. The `=` token marks where to start generating output
3. Witnesses provide "chain-of-thought" for single-token answers
4. All numbers (including multi-digit) are single tokens
5. Pattern matching uses nested list structure exactly as shown
6. Empty cycles in cycle notation are omitted (standard convention)

\end{verbatim}

\section{Fine-tuning details}
\label{sect:finetuning_hyperparameters}

We train all transformers autoregressively with the same set-up we used for training in Section \ref{sect-training}. MLPs and logistic regression are trained to predict the final answer. We include witnesses for both of the new pattern avoidance tasks (to align with PermuFormer pretraining) but we do not include witnesses for the other problems since some of these (those related to Kazhdan-Lusztig polynomials and Schubert polynomial structure constants) represent research-level tasks where natural witnesses are unknown. 

\textbf{Transformers:} Transformers were trained with AdamW using a batch size of 32 and a weight decay value of 0.01. Both versions of PermuFormer were fine-tuned using a learning rate of $1 \times 10^{-4}$. Pythia used a smaller learning rate of $5 \times 10^{-6}$ after we encountered instability issues for larger learning rates. All transformers used a linear learning-rate schedule with 100 warm-up steps. We trained the models for a maximum of 100 epochs but stopped early if accuracy failed to improve over $k$ consecutive validations (here $k$ depended on the dataset). For the pattern avoidance and mHeight datasets $k=2$, for the Schubert polynomial task $k=6$, and for the Kazhdan-Lusztig tasks $k=3$. Validation frequency depended on the dataset. For the pattern avoidance and mHeight datasets, the model was evaluated on the validation set once at the end of every epoch. For the Schubert polynomial structure constants dataset, validation was performed every 200 optimization steps. For Kazhdan-Lusztig tasks, validation occurred every 5,000 optimization steps. We report test accuracy for the model checkpoint with the best validation accuracy.

\textbf{MLPs:} Our MLP fine-tuning experiments used a different set-up. We used a batch size of 128, a learning rate of $1 \times 10^{-3}$, a weight decay of $1 \times 10^{-4}$, AdamW, and no learning rate scheduler. The early-stopping procedure for the MLPs is the same as that used for the transformers.

\textbf{Logistic regression:} We used scikit-learn’s \texttt{LogisticRegression} class to train logistic regression models with $L_2$ regularization and inverse regularization strength $C=1.0$. Binary classification tasks (like pattern detection) used the \texttt{liblinear} solver, while multiclass classification tasks used \texttt{SAGA}. The maximum number of solver iterations was 500 and the convergence tolerance was $1 \times 10^{-3}$. 

Experimentally, we found that it was beneficial to reuse existing tokens rather than introducing new tokens whose embeddings needed to be learned during fine-tuning. We provide details about this below. Dataset sizes can be found in Table \ref {table:fine-tuning size}.

\begin{itemize}

\item \textbf{Kazhdan-Lusztig polynomial coefficients:} The Kazhdan-Lusztig dataset is broken up into six different tasks: predicting the polynomial degree and predicting the coefficient on $q^k$ for $0 \leq k \leq 4$. The sizes of these sets can be found in Table \ref{table:fine-tuning size}. Differences in sizes were the result of our effort to make the datasets more balanced.

Instances were drawn from a bank of 100 million Kazhdan-Lusztig polynomials indexed by pairs of permutations from $S_9$. This bank was generated via C code from \cite{warringtonwebsite}. 

Since Bruhat order is an important concept in Kazhdan-Lusztig polynomials, we utilize tokens from this task in the training set. An example is: \texttt{n9 1linebegin [1,3,6,7,2,9,5,4,8] 1lineend leftBruhatmake 1linebegin [9,8,3,1,7,4,6,2,5] 1lineend = leftBruhatbegin 0 leftBruhatend}.
\end{itemize}

\begin{table}
    \caption{Sizes of fine-tuning datasets.}
    \label{table:fine-tuning size}
    \centering
    \begin{tabular}{llll}
        \toprule
        Fine-tuning task & Group & Training set size & Evaluation set size \\
        \midrule
        Avoids 231 & $S_9$ & 241 & 267\\
        Avoids 1342 & $S_9$ & 477 & 530\\
        mHeight & $S_9$ & 1,216 & 1,350\\
        Schubert & $S_6$ & 150,243 & 16,693\\
        KL degree & $S_9$ & 31,689 & 3,520\\
        KL coeff, $k=0$ & $S_9$ & 68,822 & 7,646\\
        KL coeff, $k=1$ & $S_9$ & 27,740 & 3,082\\
        KL coeff, $k=2$ & $S_9$ & 13,719 & 1,524\\
        KL coeff, $k=3$ & $S_9$ & 27,870 & 3,096\\
        KL coeff, $k=4$ & $S_9$ & 8,376 & 930\\
        \bottomrule
    \end{tabular}
\end{table}

\section{Interpretability experiments}

In this section we provide some details around the different interpretability experiments we performed. In all cases, hidden activations were drawn from the residual stream between blocks.

\begin{itemize}
\item \textbf{Linear probes:} All linear probes used an 80/20 train/test split. Classification probes used logistic regression with at most 1,000 iterations. Regression probes used ridge regression with $\alpha=1$.

\item \textbf{Visualizations:} Visualization was generated by sampling 1,000 different permutations which were then expressed in all six encodings. We then injected them into specific task prompts. UMAP used \texttt{n\_neighbors} equal to 5 and \texttt{min\_dist} equal to 0.1.

\item \textbf{CKA:} We used linear CKA with 100 examples drawn from the bank of 1,000 permutations listed above. These same permutations were used across the experiments. Activations were drawn from the 6th block at the first generated token.
\end{itemize}



\end{document}